\documentclass[twoside]{article}

\usepackage[accepted]{aistats2022}

\usepackage[round]{natbib}

\usepackage{graphicx}
\usepackage{enumitem}
\usepackage{amsmath,amssymb,amsfonts,amsthm}

\usepackage{graphicx}
\usepackage{amsmath,amsthm,amssymb}
\DeclareMathOperator*{\argmax}{arg\,max}

\usepackage{xcolor}

\usepackage{verbatim}
\newcommand{\cmmnt}[1]{}
\usepackage{enumitem}
\usepackage{xspace} %

\usepackage{natbib}
\setcitestyle{round}

\newtheorem{theorem}{Theorem} 
 
\newtheorem{proposition}[theorem]{Proposition}

\newtheorem{definition}[theorem]{Definition}

\newtheorem{conjecture}[theorem]{Conjecture}

\newcommand{\Algorithm}{\mbox{P2-EXP4}\xspace} %
\DeclareMathOperator{\Reg}{Reg}

\begin{document}

\runningtitle{A Bandit Model for Human-Machine Decision Making with Private Information and Opacity}

\twocolumn[

\aistatstitle{A Bandit Model for Human-Machine Decision Making\\ with Private Information and Opacity}

\aistatsauthor{ Sebastian Bordt \And Ulrike von Luxburg}

\aistatsaddress{University of T{\"u}bingen\\Max Planck Institute for Intelligent Systems\\
T{\"u}bingen, Germany \And University of T{\"u}bingen\\Max Planck Institute for Intelligent Systems\\
T{\"u}bingen, Germany} ]

\begin{abstract}
Applications of machine learning inform human decision makers in a broad range of tasks. The resulting problem is usually formulated in terms of a single decision maker. We argue that it should rather be described as a two-player learning problem where one player is the machine and the other the human. While both players try to optimize the final decision, the setup is often characterized by (1) the presence of private information and (2) opacity, that is imperfect understanding between the decision makers. We prove that both properties can complicate decision making considerably. A lower bound quantifies the worst-case hardness of optimally advising a decision maker who is opaque or has access to private information. An upper bound shows that a simple coordination strategy is nearly minimax optimal. More efficient learning is possible under certain assumptions on the problem, for example that both players learn to take actions independently. Such assumptions are implicit in existing literature, for example in medical applications of machine learning, but have not been described or justified theoretically.
\end{abstract}

\section{Introduction}
\label{sec:introduction}

The number of applications where machine learning informs human decision makers is steadily growing \citep{reserve2007report,angwin2016machine,tonekaboni2018prediction}. In this work, we argue for a specific perspective on machine learning systems that inform human decision makers: We want to understand them as attempts to solve joint human-machine decision making problems where both sides have to learn to act optimally. This perspective will help to understand limitations and potential pitfalls of such systems. We believe that this is an important step towards robust and reliable systems \citep{rahwan2019machine}.

Our motivation is the growing number of applications where machine learning advises human decision makers. For example:
\begin{enumerate}
    \item[(1)] The COMPAS program that assists judges during criminal trials \citep{angwin2016machine}. The program provides a risk assessment score for defendants in criminal law. Judges then use this score, among others, to decide whether a defendant should await trial at home or in jail, and to determine the length of prison sentences \citep{kleinberg2018human,forrest2021machines}.
    
    \item[(2)] Cardiac arrest and other forms of adverse event prediction. In medicine and beyond, it can be of great value to know when adverse events such as cardiac arrest are likely to occur \citep{tonekaboni2018prediction,8844833,baker2020continuous}. This can often be predicted based on a limited amount of information. Computer programs alert doctors when a  patient’s  condition  is  likely  to  become critical. Doctors respond with a treatment adapted to the patient's condition, which may include ignoring the alert. 
    
    \item[(3)] Diabetic retinopathy detection \citep{raghu2019algorithmic}. Deep learning has shown great capabilities to detect diabetic retinopathy in pictures of the eye. This has led to  computer programs that inform doctors by assigning scores to to images. Doctors incorporate these scores into their decision making \citep{beede2020human}.
\end{enumerate}

\begin{figure*}[t]
\begin{center}
\includegraphics[width=0.92\textwidth]{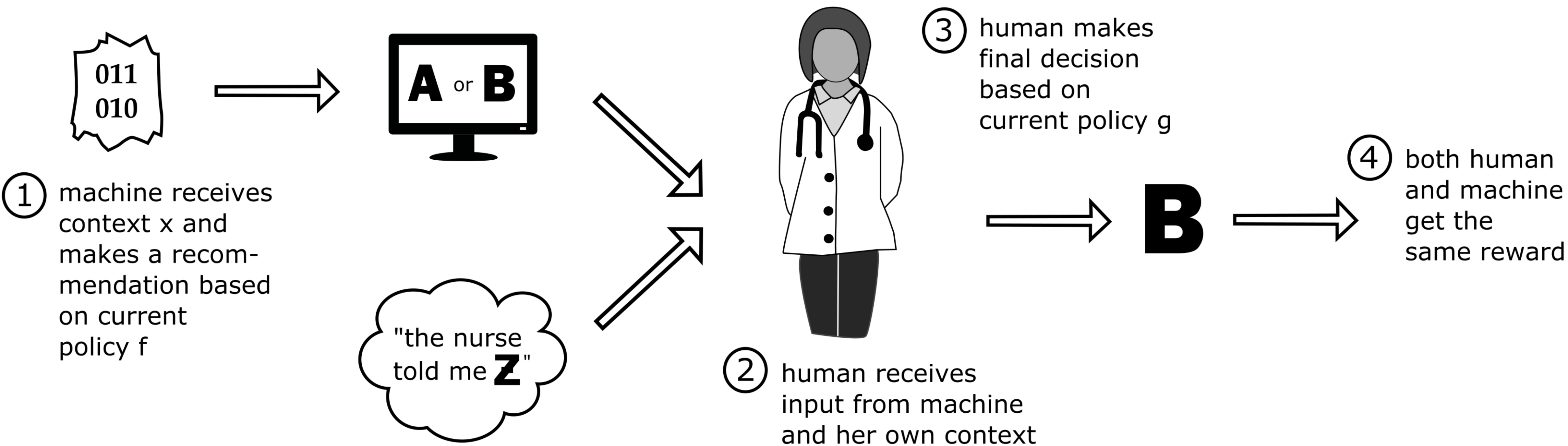}    
\end{center}
\caption{Illustrated application of our model: The computer is advising a human doctor. The computer recommends to perform action A or B. After consulting her additional private information, the human finally decides to perform action B.}
\label{fig:page_2_figure}
\end{figure*}

In all three examples, machine learning provides advice, but final decisions are left to the human. Moreover, human decision makers base their decisions on additional {\it private information} that is unavailable to the machine. In the COMPAS example, the judge obtains additional information from the trail and the interaction with the defendant,  attorney and prosecutor \citep{lakkaraju2017selective}. In the medical example, private information might consist of non-digitized parts of the patient's medical history, or diseases that run in the family of the patient \citep{goldenberg2019health}. Even in the diabetic retinopathy example, the final treatment decision is typically based on more information than just the picture of the eye.

In addition to the presence of private information, it has long been argued that human-machine cooperation is hampered by a certain degree of {\it opacity} \citep{sep-science-big-data}. Indeed, despite a lot of work on explainable machine learning, computers cannot explain their decisions to humans the way other humans can, and computers cannot really understand free-form human explanations. 

How can we design computer programs that optimally advise human decision makers in tasks such as (1)-(3)? What does ``optimally'' even mean in these contexts? To provide precise answers to these questions, we propose a {\it contextual bandit model with two players} that aims to capture the most important properties of the above decision problems. 

The two players in our model, who we refer to as ``the human'' and ``the machine'', interact according to the following protocol (illustrated in Figure \ref{fig:page_2_figure}). In every round, the first player (the machine) receives private contextual information and makes a recommendation to the second player (the human). This recommendation can be a suggested action, but it can also be a confidence region, a colorfully highlighted image or any other summary of the received context. In the COMPAS example, the recommendation is the risk assessment score. Given the recommendation and her own private contextual information, the human finally decides on an action. Conditional on context and the chosen action, a reward signal is obtained. Action and reward are observed by both players, and they share the same goal: to maximize the obtained rewards. 

We endow each of the players with a finite set of decision rules or policies, which is the simplest possible learning setting. The goal is to minimize the minimax regret with respect to the two decision rules that work best together. We first analyze the case where the human does not attempt to learn (Section \ref{sec:lower_bound}). However, we believe that the fact that human decision makers have to learn how to ``interpret'' machine recommendations is a crucial aspect of human-machine decision making. In cardiac arrest prediction, for example, doctors have reported to learn over time how to interpret machine alerts and integrate them into existing clinical practice. We therefore also consider the problem where human and machine {\it both} have to learn (Section~\ref{sec:upper_bound}). 

The main objective of this paper is to gain a theoretical understanding of an emerging number of human-machine decision making problems, such as (1)-(3). By considering the interaction between two abstract decision makers in the presence of private information and opacity, we aim to provide a general analysis of the potential and limitations of human-machine decision making. While our main intention is to set a theoretical baseline for more applied work in human-machine interaction, we also hope that our proposed model and newly introduced problems will spark theoreticians interest into various aspects of the human-machine learning problem. Our main contributions are the following.\\[12pt]

\begin{itemize}
\item  We prove that private information and opacity significantly impact the hardness of two-player decision making. Private information and opacity each lead to a worst-case {\it lower bound} of order $\sqrt{TN_1}$ (Theorem \ref{thm:lower_bound}). Here $N_1$ is the number of policies of the first player. Without private information and opacity, the two players can obtain an efficient expected regret of $\sqrt{2TK\ln(N_1N_2)}$ (Proposition \ref{prop:no_private_no_opacity}). Here $N_2$ is the number of policies of the second player.

\item We show that a simple coordination device -- telling the machine which policy to use -- allows the human to learn efficiently. Specifically, the \Algorithm algorithm allows to {\it upper bound} the expected minimax regret by $\sqrt{2TK N_1\ln(N_1 N_2)}$ (Theorem \ref{thm:upper_bound}), also in the presence of private information and opacity.  

\item We derive a criterion  -- policy space independence -- that allows to learn with an expected regret of $\sqrt{8T\max\{K,|\mathcal{R}|\}\ln(\max\{N_1,N_2\})}$ (Theorem \ref{thm:policy_space_independence}). Here $|\mathcal{R}|$ is the number of possible machine recommendations. If policy space independence holds and $|\mathcal{R}|$ is small, the two players can learn efficiently.

\item In Sections \ref{sec:beyond_worst_case} and \ref{sec:treatment_recommendations}, we show that various approaches in the literature can be better understood within the context of our model. In particular, policy space independence is implicit in much of the existing literature. The peculiar case of treatment recommendations is left as Conjecture \ref{thm:conjecture}.
\end{itemize}

\section{Our model: The computer reports to the human, who then decides}
\label{sec:prelimiaries}

Formally, our model is a contextual bandit model with two players, depicted in Figure \ref{fig:model}. In round $t=1,\dots,T$, Player 1 (the machine) first observes context $x_t\in\mathcal{X}$. Player 1 then chooses a recommendation $r_t\in\mathcal{R}$, potentially at random. Here, $\mathcal{R}$ is the space of all possible recommendations that the first player can make. Next, Player 2 (the human) observes context $z_t\in\mathcal{Z}$ and the chosen recommendation $r_t$. Player 2 then, potentially at random, chooses an action $a_t\in A$. This action is revealed to both players, and they receive a reward signal $y_t\in[0,1]$. Here $\mathcal{X}$ and $\mathcal{Z}$ are arbitrary spaces of private contexts (one for each player), and $A=\{1,\dots,K\}$ is a finite set of $K$ actions.

\subsection{Formal setup} \label{sec:formal_setup}

Both players are endowed with a finite set of policies. Their common goal is to 
take optimal actions. Let $\Pi_1\subseteq \mathcal{R}^\mathcal{X}$ be a finite set of policies for the first player, and $\Pi_2\subseteq A^{\mathcal{R}\times \mathcal{Z}}$ a finite set of policies for the second player. Given two policies $f\in\Pi_1$ and $g\in\Pi_2$, we obtain the resulting joint policy $\pi(x,z)=g(f(x),z)$. This joint policy is a complete decision rule for the problem, translating context into actions. Let $\Pi=\Pi_2\times\Pi_1$ be the space of all combinations of policies that the two players can possibly realize. For a tuple $\pi=(g,f)\in\Pi$, we slightly abuse notation and write $\pi(x,z)=g(f(x),z)$ to refer to the corresponding joint policy.\footnote{Depending on $\Pi_1$ and $\Pi_2$, different tuples $(g,f)$ can give rise to the same policy $\pi:\mathcal{X}\times\mathcal{Z}\to A$.} Moreover, we denote the number of policies $N_1=|\Pi_1|$ and $N_2=|\Pi_2|$. We have $N=|\Pi|=N_1 N_2$.

An algorithm for the two players is a pair $A=(A_1,A_2)$. Here $A_1=(A_{1,t})_{t=1}^T$ and $A_2=(A_{2,t})_{t=1}^T$ are two collections of measurable functions that specify the decision rules of both players at all points in time. The domains of these functions specify which variables are observable to which player at what time. Thus, $A_{1,t}$ is a function of $x_1,\dots,x_t$, whereas $A_{2,t}$ is a function of $r_1,\dots,r_t$ and $z_1,\dots,z_t$. The details of this can be found in Supplement \ref{apx:defintions}.

Let $\mathcal{D}$ be a probability distribution over $\mathcal{X}\times \mathcal{Z}\times [0,1]^A$. We consider an i.i.d.\ contextual bandit model where tuples $(x_t,z_t,Y_{t})$  are i.i.d.\ draws from $\mathcal{D}$. Let $Y(\pi)=\mathbb{E}_{(x,z,Y)\sim\mathcal{D}}\left[Y(\pi(x,z))\right]$ be the expected reward of a joint policy $\pi$. Let $\pi^\star\in\argmax_{\pi\in\Pi}Y(\pi)$ be a policy combination that maximizes the expected reward. The expected regret after $T$ rounds is given by $\Reg_{T}=\mathbb{E}\left[T\,Y(\pi^\star)-\sum_{t=1}^T Y_t(a_t)\right]$, where the expectation is over $\mathcal{D}$ and the randomly selected actions and recommendations. The central quantity of analysis is the minimax regret, given by $R_T =\inf_{A}\sup_{\mathcal{D}}\sup_{|\Pi_1|=N_1}\sup_{|\Pi_2|=N_2}\,\Reg_T$.

\begin{figure}[t]\centering
\fbox{\parbox{0.46\textwidth}{In round $t=1,...,T$
\begin{itemize} 
	\item[1.] Context $x_{t}\in\mathcal{X}$ is revealed to Player 1
	\item[2.] Player 1 decides on a recommendation $r_t\in\mathcal{R}$
	\item[3.] Context $z_{t}\in\mathcal{Z}$ and recommendation $r_t$ are\\ revealed to Player 2
	\item[4.] Player 2 decides on an action $a_t\in A$
	\item[5.] Reward $y_t\in[0,1]$ and action $a_t$ are revealed\\ to both players
\end{itemize}}}
\caption{Interaction in our contextual bandit model.}
\label{fig:model}
\end{figure}

\subsection{First thoughts and discussion of modelling assumptions} \label{sec:first_thoughts}

\textbf{Private context.} This is our approach to model private information. In real-world decision making problems such as (1)-(3), humans often have access to information that is not available to any algorithm. A reason for this might be that some information, such a detailed health record, is not yet available in electronic form. However, we also believe that in many of the tasks where machine learning is increasingly being deployed at, formulating all relevant aspects as inputs to an algorithm is impossible. This is because machine learning is increasingly being deployed in social contexts where researchers have long accepted the fact that its impossible to exhaustively collect all relevant variables \citep{angrist2008mostly}. Our model also allows for private contextual information of the machine. In the medical domain, an algorithm might have access to a patient's genome data, which could never be entirely surveyed by a human. Unobserved variables might also occur in unexpected situations, such as when both decision makers coordinate a decision based on the same image. Here algorithms have been shown to rely on high-frequency patterns that are imperceptible to humans \citep{NEURIPS2019_e2c420d9, makino2020differences}.

\textbf{Private policy spaces.} We model opacity by keeping knowledge about the policy spaces to the respective players. Intuitively, this means that the players cannot deliberate about what happened: The machine does not know which actions the human would have chosen had it chosen a different recommendation. Similarly, the human does not know which recommendations the machine considered but decided against. While policy spaces are private, we place no restrictions on the algorithms that both players might run.

\textbf{The space of recommendations.} The space of recommendations $\mathcal{R}$ is the interface by which the first player can transmit information to the second player \citep{Goodrich2007}. For the first player, it plays the role of an action space (providing a recommendation is the action that the first player takes). For the second player, it resembles additional contextual information. In the analysis, will turn out to be useful to restrict the size of the space of recommendations (Section \ref{sec:beyond_worst_case}). A large space of recommendations allows the machine to provide the human with rich contextual information. This includes the scenario where the machine attempts to ``explain'' predictions in some rich space. A concrete example of this would be when the machine provides a saliency map \citep{simonyan2013deep, selvaraju2017grad}. In contrast, a small space of recommendations allows the machine to suggest concrete actions, or to raise an alert. A priori, it seems unclear which of these two approaches will be more useful. On one hand, we might want the machine to provide the human with as much information as possible. On the other hand, it might be more efficient if the machine directly suggests which actions to perform. In Sections 3-5, we remain agnostic about the nature of the space of recommendations. The special case of treatment recommendations is discussed in Section \ref{sec:treatment_recommendations}.

\textbf{What makes the model difficult?} For both players, the difficulty arises from the fact that contextual information and policy space of the other player are unknown. This gives rise to a {\it coordination problem}. Each player would like to find the optimal policy that works best in combination with the strategy chosen by the other player. This is difficult because knowledge about the other player's decision problem is limited.

\textbf{Online learning.} Our model is an online learning model. This allows us to study the process by which the two decision makers coordinate and arrive at decisions. In practice, an algorithm would always be trained on a historical dataset before it starts to interact with a human decision maker. However, if we continuously gather data in order to retrain and improve our algorithm, we are implicitly engaging in an online learning procedure. We are directly considering an online learning model since this allows us to study the principal limitations and possibilities of various approaches. For more details on online and repeated supervised learning we refer the reader to Supplement~\ref{apx:online_learning}.

\textbf{Worst-case analysis.} Intuitively, a strategy of the two players might work well for some decision problems and fail for others. Considering the minimax regret means that we would like to find guarantees that can be achieved under {\it all possible circumstances}. That said, it is interesting to ask how much better the two players can do if we assume that the decision problem is 'benign' -- a question that we turn to in Section \ref{sec:beyond_worst_case}.

\section{Two baselines for the expected regret}

How well can we expect the two players to coordinate, and what are the consequences of private information and opacity for two-player decision making? To provide answers to these questions, we are first going to consider our model {\it without} private information and opacity. No private information means that $\mathcal{X}=\mathcal{Z}$ and $x_t=z_t$ for all $t$. No opacity means that the algorithm of the first player is also a function of the policy space of the second player and vice-versa. 

\begin{proposition}\textbf{(Regret without private information and opacity)}\label{prop:no_private_no_opacity} 
Without private information and opacity, the two players can obtain an expected regret of
\begin{equation*}
\sqrt{2TK \ln (N_1 N_2)}.
\end{equation*}
\end{proposition}

All proofs are deferred to the Supplement. The regret bound in Proposition \ref{prop:no_private_no_opacity} is as good as we can expect at all.\footnote{For adversarial contextual bandits, the bound $\sqrt{TK\ln(N)}$ has been shown to be tight up to a factor of $\ln K$ by \citet{seldin2016lower}. Note that we are concerned with statistical optimality and set computational concerns aside.} It is the same regret that would be achieved by a hypothetical single decision maker who had access to all contextual information and both policy spaces, using EXP4 \citep{auer2002nonstochastic,lattimore2019bandit}. Proposition \ref{prop:no_private_no_opacity} demonstrates that our hardness result (Theorem \ref{thm:lower_bound}) is a consequence of private information and opacity, and not due to the way in which the two players interact in our model. 

A second baseline is given by the coordination strategy where players naively try all policy combinations. This strategy also works {\it with} private information and opacity. Using the MOSS algorithm by  \cite{audibert2010regret}:

\begin{proposition}\textbf{(Naively trying all policy combinations)}\label{proposition1} 
By treating the policies in $\Pi$ as different arms of a stochastic bandit, we obtain
\begin{equation*}
R_T\leq O\left(\sqrt{TN_1N_2}\right).
\end{equation*}
\end{proposition}

Under private information and opacity, can the two players do better than what is suggested by Proposition \ref{proposition1}? The question becomes whether it is possible to {\it move $N_1$ and $N_2$ inside the logarithm}, at the expense of a factor of $K$. Why is this important? A regret bound of order $\sqrt{N}$ means that the policy space is not dealt with efficiently. It corresponds to systematic trial and error on every single policy. Quite to the contrary, a regret bound of order $\sqrt{\ln N}$ means that the decision maker can compare many policies simultaneously. It prepares the way to deal with infinite policy spaces and learn rich function classes  \citep{Beygelzimer2010ContextualBA}. 

\section{A lower bound for optimal algorithmic advice} \label{sec:lower_bound}

Before we turn to the full problem where human and machine both have to learn, we focus on the problem of the machine. That is we assume that the human does not have to learn how to interpret machine recommendations. This is a significant simplification, but the result will be instructive. We are going to show that private information and opacity {\it each} lead to a lower of order $\sqrt{T N_1}$.

Formally, we assume that the second player follows a fixed decision rule that deterministically translates recommendations $r_t$ and contextual information $z_t$ into actions. The first player has $N_1$ different policies and wants to learn the best one. How difficult is this learning problem? Note that we do not place any restrictions on the space of recommendations $\mathcal{R}$. However, the ultimate number of actions $K$ is small. Can the first player make use of this fact and solve the problem efficiently? In the presence of private information or opacity, this is not the case.

\begin{theorem}\textbf{(Lower bound in the number of policies of the first player)}\label{thm:lower_bound}
Assume that Player~2 only plays actions that are suggested by policies in $\Pi_2$. Let $N_2=1$ and $K=2$. There exists a universal constant $c>0$ such that
\begin{equation*}
R_T\geq c\sqrt{TN_1}.
\end{equation*}
\end{theorem} 

The lower bound in Theorem \ref{thm:lower_bound} is as strong as it can possibly be. It shows that the first player has to solve a bandit problem that depends {\it not} on the number of actions $K$, but on the number of policies $N_1$. 

Supplement \ref{apx:proof_lower} contains two proofs of Theorem \ref{thm:lower_bound}. The first proof constructs problem instances with private information but without opacity, and the second proof constructs problem instances with opacity but without private information.\footnote{From a theoretical perspective it might not be surprising that private information and opacity have the same consequences. Ultimately, what matters are the expert predictions that result from the interaction of context and policy \citep{cesa2006prediction}.} %
In conclusion, both private information and opacity can significantly impact on the hardness of two-player decision making. 

{\bf Remark 1.} The reader might be worried by the fact that we fixed Player 2. Indeed, even if the policy space of the second player consists of a single decision rule, it might be optimal to deviate in order to facilitate coordination. To alleviate such concerns, Supplement \ref{apx:lower_assumption_construction} presents a problem class that allows the second player to choose actions arbitrarily. We discuss the general issue of encoding information about the policy spaces in actions and recommendations in Supplement \ref{apx:implicit_coordination}.

\section{Efficient learning for a human who controls the machine}
\label{sec:upper_bound}

\begin{figure}[t]
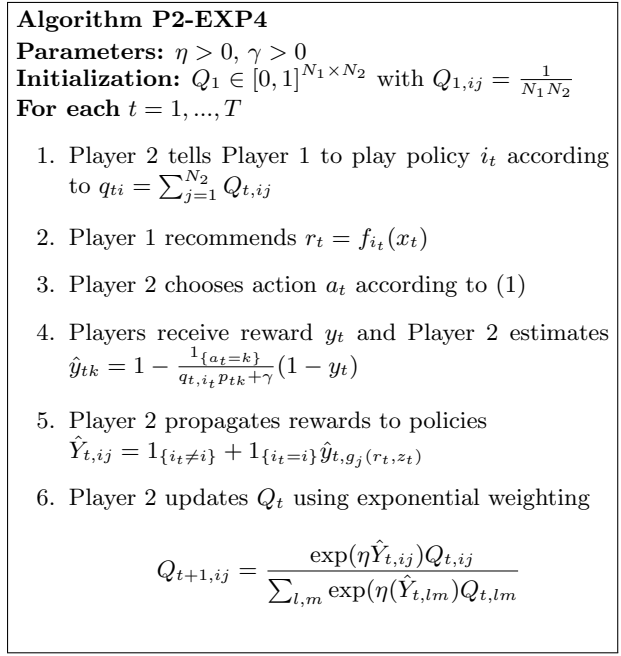
\centering
\fbox{\small\parbox{0.46\textwidth}{\textbf{Algorithm \Algorithm}\\[2pt]
\textbf{Parameters:} $\eta>0$, $\gamma>0$\\
\textbf{Initialization:} $Q_{1}\in [0,1]^{N_1\times N_2}$ with $Q_{1,ij}=\frac{1}{N_1N_2}$\\
\textbf{For each} $t=1,...,T$
\begin{itemize} 
	\item[1.] Player 2 tells Player 1 to play policy $i_t$ according to $q_{ti}=\sum_{j=1}^{N_2} Q_{t,ij}$
	\item[2.] Player 1 recommends $r_t=f_{i_t}(x_t)$
    \item[3.] Player 2 chooses action $a_t$ according to (\ref{eq:alg1_pt})
	\item[4.] Players receive reward $y_t$ and Player 2 estimates $\hat y_{tk}=1-\frac{1_{\{a_t=k\}}}{q_{t,i_t}p_{tk}+\gamma}(1-y_t)$
    \item[5.] Player 2 propagates rewards to policies\\ $\hat Y_{t,ij}=1_{\{i_t\neq i\}}+1_{\{i_t=i\}}\hat y_{t,g_j(r_t,z_t)}$
    \item[6.] Player 2 updates $Q_t$ using exponential weighting\begin{equation*}
    Q_{t+1,ij}=\frac{\exp(\eta \hat Y_{t,ij}) Q_{t,ij}}{\sum_{l,m}\exp(\eta (\hat Y_{t,lm})Q_{t,lm}}
    \end{equation*}
    \end{itemize}}}
\caption{The \Algorithm algorithm allows the second player to explore efficiently.}
\label{fig:alg1}
\end{figure}

We now consider the full problem of human-machine learning where the human learns how to interpret machine recommendations. Intuitively, the human tries to figure out how to act on machine advice. At the same time, the machine tries to determine how to advise the human. How should the two players coordinate? Is it possible that both explore simultaneously? 

From Theorem \ref{thm:lower_bound} in the previous section, we already know that the learning problem of the machine is hard, even if the human sticks to a single policy. We are now going to show that a simple coordination device allows the human to explore efficiently: {\it We allow the human to tell the machine which policy to use}. Intuitively, we can perceive the $N_1$ different policies of the machine as different computer programs. The human tries to learn which of these computer programs to use. In doing so, the human can explore with exponential weighting, but only on its own policy space, and not on the policy space of the machine. This idea is formalized in the \Algorithm (Player 2-EXP4) algorithm, depicted in Figure \ref{fig:alg1}. Theorem  \ref{thm:upper_bound} shows that \Algorithm nearly allows to match the lower bound in Theorem \ref{thm:lower_bound}.

\begin{theorem}\textbf{(Logarithmic regret in the number of policies of the second player)} \label{thm:upper_bound} 
The \Algorithm algorithm with $\eta=\sqrt{2\log(N_1N_2)/(TKN_1)}$ and $\gamma=0$ satisfies
\begin{equation*}
R_T\leq \sqrt{2TK N_1\ln(N_1 N_2)}.
\end{equation*}
\end{theorem}

The proof of Theorem \ref{thm:upper_bound} is in Supplement \ref{apx:proof_upper}. We now describe the algorithm. Player 2 maintains a probability distribution $Q_t$ over the space of all policies $\Pi$. In every round $t$, Player 2 first chooses a policy $f_{i_t}$ for Player 1 by drawing $i_t$ from the marginal distribution of $Q_t$ over $\Pi_1$. After obtaining a recommendation $r_t$ and context $z_t$, Player 2 draws $a_t$ according to the induced probability distribution over actions 
\begin{equation}\label{eq:alg1_pt}
\mathbb{P}(a_t=k)=p_{tk}\quad\text{with}\quad p_{tk}=\frac{\sum_{j=1}^{N_2} Q_{t,i_t,j}1_{\{g_{j}(r_t, z_t)=k\}}}{\sum_{j=1}^{N_2} Q_{t,i_t,j}}.   
\end{equation} With the reward signal $y_t$, Player 2 computes importance-weighted reward estimates for all policies and then uses an exponential weighting scheme to update $Q_t$. 

{\bf Remark 2.} The proof of Theorem \ref{thm:upper_bound} relies on the fact that the updates performed by \Algorithm are equivalent to the updates performed by EXP4 on a related bandit problem with $K N_1$ actions. As a consequence, all results for EXP4 carry over to \Algorithm. In particular, for $\gamma>0$, \Algorithm is a variant of EXP4-IX \citep{neu2015.explore.no.more}. This implies that \Algorithm enjoys high-probability regret guarantees. Furthermore, Theorem~\ref{thm:upper_bound} also holds when contexts and payoffs are determined by an adversary.

\section{Efficient learning for the machine, subject to further assumptions}
\label{sec:beyond_worst_case}

We now discuss additional assumptions on the structure of the problem that allow for more efficient learning. The first idea is to restrict the size of the space of recommendations $\mathcal{R}$. If the machine directly recommends actions, for example, we have $\mathcal{R}=A$. The second idea is to resolve the coordination problem. This can be done via an assumption on the function spaces of both players that we term {\it policy space independence}. While policy space independence is an abstract criterion, we outline a number of practical examples where it is satisfied.

This section also relates our work to a number of recently proposed techniques for human-machine interaction \citep{madras2018predict, raghu2019algorithmic, wilder2020learning}. We will show that policy space independence is implicit in much of the existing literature on human-machine decision making.

\subsection{Policy space independence}

We now give an abstract condition that {\it resolves the coordination problem} and allows both players to learn independently. It is an assumption on the policy spaces. The rationale is that assumptions on the policy spaces can implicitly define how human and machine interact.

\begin{definition}[\bf Policy space independence] \label{def:policy_space_independence}
We say that the two policy spaces $\Pi_1$ and $\Pi_2$ are independent with respect to $\mathcal{D}$ if, for all $f_1,f_2\in\Pi_1$ and all $g_1,g_2\in\Pi_2$,
\begin{equation*}
\begin{split}
 Y\big(g_1(f_1(x),z)\big)&-Y\big(g_1(f_2(x),z)\big)\\=\,\,Y\big(g_2(f_1(x),z)\big))&-Y\big(g_2(f_2(x),z)\big).      
\end{split}
\end{equation*}
\end{definition}

Intuitively, whether policy $f_1$ performs better than policy $f_2$ does not depend on the policy chosen by the second player. Similarly, whether policy $g_1$ performs better than policy $g_2$ does not depend on the policy chosen by the first player. Hence, the learning problems of both players are decoupled. The following theorem shows that policy space independence allows to efficiently learn both policy spaces.

\begin{theorem}[\bf Logarithmic regret under policy space independence] \label{thm:policy_space_independence}
Under policy space independence, if both players explore independently using EXP4,
\begin{equation*}
R_T\leq \sqrt{8T\max\{K,|\mathcal{R}|\}\ln(\max\{N_1,N_2\})}.
\end{equation*}
\end{theorem}

The proof of Theorem \ref{thm:policy_space_independence} is in Supplement \ref{apx:proof_policy_space_independence}. In contrast to Theorem \ref{thm:upper_bound}, both $N_1$ and $N_2$ appear inside the logarithm. This is at the expense of a factor $|\mathcal{R}|$. 

\subsection{Allocating decisions between human and machine} \label{sec:fixed_allocation_rules}

If $|\mathcal{R}|$ is small and policy space independence holds, the two players can obtain an efficient expected regret (Theorem \ref{thm:policy_space_independence}). But what does this amount to in practice? First note that we can constrain the policy space of the human by specifying rules for how to interact with the machine. For example: ``If the machine depicts 'action $a$', then perform action $a$''. This leads to the following example: Policy space independence holds when there exists a fixed rule that allocates every decision to {\it either} the human {\it or} the machine. In medical applications, this would mean that there exists some procedure that determines whether a given case should be decided by the doctor or the machine. For diabetic retinopathy detection, such a procedure was recently proposed by \citet{raghu2019algorithmic}, who also demonstrate that the approach can lead to substantial benefits in practice. In our model, the rule can be any predicate $P(z)$, that is the human decides who decides. It can also be any predicate $P(x)$, that is the machine decides who decides. Importantly, in order to satisfy policy space independence, the rule cannot be learned while the decision makers learn themselves. We formally show in Supplement~\ref{apx:fixed_allocation_rules} how fixed rules that allocate decisions result in policy space independence.

\subsection{Learning to defer}

Another example of policy space independence is given by learning to defer \citep{madras2018predict, mozannar2020consistent}. Learning to defer is characterized by two assumptions. First, the human is a fixed decision maker who does not learn. Second, the space of recommendations is given by $\mathcal{R}=\mathcal{A}\cup\{\texttt{D}\}$, where \texttt{D} denotes that the decision is deferred to the human. As can be seen from Definition \ref{def:policy_space_independence}, fixing any of the two decision makers always results in policy space independence. According to Theorem \ref{thm:policy_space_independence}, the regret of learning to defer is thus bounded by $\sqrt{8T(K+1)\ln(N_1)}$.\footnote{For $N_2=1$, the constant could be improved to 2.}

\subsection{Other approaches} 

With some notable exceptions \citep{hilgard2019learning}, the literature on human-machine decision making often relies on assumptions similar to fixed rules that allocate decisions and learning to defer \citep{de2020regression, de2020classification}. It is usually assumed that the human is a fixed decision maker whose performance on the given task can be queried or deferred to \citep{wilder2020learning, pradier2021preferential}. Specifically, the human does not have to learn how to interact with the machine. Moreover, machine recommendations usually equal actions, with some room for special recommendations in order to involve the human. Viewed through the lens of our model, all of these approaches satisfy policy space independence. In light of Theorem \ref{thm:policy_space_independence}, they all allow for efficient learning. 

\section{How difficult are treatment recommendations?}
\label{sec:treatment_recommendations}

In the last section, we have seen that the learning problem of the machine can be simplified by (1) choosing $\mathcal{R}=\mathcal{A}$ and (2) fixing the human decision maker. In Section~\ref{sec:lower_bound}, we have seen that the condition $\mathcal{R}=\mathcal{A}$ is crucial (after all, the lower bound was derived for a fixed human decision maker). But is it equally necessary to choose $|\Pi_2|=1$? This is interesting because $\mathcal{R}=\mathcal{A}$ is satisfied, among others, in screening scenarios. These are the binary classification problems studied in the literature on fairness and machine learning \citep{kleinberg2018discrimination, barocas-hardt-narayanan}. Here a decision problem might be whether to give a loan or to admit a student to a university. It is often argued that such machine suggestion should still be reviewed by humans \citep{de2020case}. 

In our model, binary predictions that are reviewed by humans correspond to $\mathcal{R}=A=\{0,1\}$ and $|\Pi_2|>1$ (assuming that the human learns when to override machine predictions). If either $N_1=1$ or $N_2=1$, EXP4 allows to bound the expected regret by $\sqrt{4T\ln(N_2)}$ and $\sqrt{4T\ln(N_1)}$, respectively. Therefore, consider the corner case $N_1=N_2$. If we assume that the second player can tell the first player which policy to use, \Algorithm allows to bound the expected regret by $\sqrt{8TN_1\ln(N_1)}$. We conjecture that this is tight up to a constant factor, i.e. that treatment recommendations are difficult.

\begin{conjecture}{\textbf{(Lower bound in the number of policies if $\mathcal{R}$ and $A$ are small)}} \label{thm:conjecture}
Let $\mathcal{R}=A=\{0,1\}$ and $N_1=N_2$. We conjecture that there exists a universal constant $c>0$ such that
\begin{equation*}
R_T\geq c\sqrt{TN_1\ln N_1}.
\end{equation*}
\end{conjecture}

Supplement \ref{apx:conjecture} details a problem instance that we believe to be worst-case.\footnote{The reader might wonder whether interaction terms between $K$ and $\mathcal{R}$ appear in any bound. Beyond the special regime $\mathcal{R}=A=\{0,1\}$ and $N_1=N_2$, this might well be the case. }

\section{Related Literature}

Researchers have long asked how humans can interact with computers and robots \citep{sheridan1978human, Goodrich2007, parasuraman2000model}. In machine learning, researchers increasingly study how humans and automated decision making systems can interact \citep{tonekaboni2019clinicians, NIPS2019_8760, lucic2020does, de2020case}. A number of recent works have argued that joint human-machine decision making can outperform a single human or a single machine \citep{lakhani2017deep, raghu2019algorithmic, Patel_2019}. Human-computer interaction and the social sciences study the different ways in which machine recommendations can influence and alter human decisions \citep{dietvorst2015algorithm, green2019principles}.

Multi-player multi-armed bandits \citep{kalathil2014decentralized, boursier2019sic,NEURIPS2019_85353d3b}, economic game theory \citep{mas1995microeconomic,von2007theory} and combinations thereof \citep{sankararaman2021dominate} also study the interaction between multiple players. However, models in economic game theory are {\it competitive}, and the {\it cooperative} models in multi-player multi-armed bandits, often inspired by applications in wireless networks \citep{avner2016multi}, are {\it symmetric}. In contrast, interaction our model is cooperative and {\it asymmetric} -- only the second player decides on a payoff-relevant action. Insofar as implicit communication between the two players is concerned, our work probably relates most closely to \citet{bubeck2020non}, who study implicit communication in a symmetric collision problem (compare also Supplement~\ref{apx:implicit_coordination}).

\section{Discussion} \label{sec:discussion}

{\bf The consequences of private information and opacity.} We have shown that private information and opacity can have a significant effect on human-machine decision making. In the worst-case, the machine cannot advance beyond simple trial and error on a small number of policies (Theorem \ref{thm:lower_bound}). Does this imply that we can never obtain good results in general human-machine decision making problems where we cannot make plausible assumptions on the presence of private information and opacity? Not necessarily. It does, however, imply that we need good priors for what comprises successful human-machine cooperation on a given task. Note that in practice, researchers often obtain a small number of candidate machine policies from historical data, then evaluate which one works best with human decision makers \citep{sayres2019using, tschandl2020human}. This approach is closely related to running the \Algorithm algorithm: The policy space of the machine consists of the candidate decision rules that were obtained form historical data. In the absence of further assumptions about the problem, we show this approach to be essentially minimax optimal. In some applications, it might be relatively easy to come up with good machine policies. There are, however, also problems where it is hard say how the machine should best inform the human. Consider the example where the machine informs the human about an image: While there have been many empirically successful attempts at such problems, there is still a big debate about post-hoc explainability methods, what properties they should have, and whether they should be used at all \citep{adebayo2018sanity,rudin2019stop}. 

{\bf Different modalities of human-machine decision making.} We have seen in Section \ref{sec:beyond_worst_case} that our model possesses sufficient generality to analyze a wide array of interaction protocols between humans and machines. Of course, there are many different settings of human-machine decision making, and our model can only serve as first step towards a formal analysis. From a theoretical perspective, it remains an interesting open question whether there are weaker assumptions than policy space independence that allow for efficient learning (Theorem \ref{thm:policy_space_independence}). One might also ask whether distributional assumptions that restrict the influence of unobserved variables on the outcome can result in improved bounds.  From a practitioner's point of view, the most important question is which assumptions are plausibly satisfied in applications.

{\bf Prediction problems.} In many decision support systems,  machine learning is merely used to solve a specific prediction or classification problem, whose outcome is then transferred to the human. Examples are scores to predict criminal recidivism, cardiac arrest and severity of diabetic retinopathy. While such an approach is a straightforward way of human-machine interaction, nothing guarantees that the approach will be successful. For example, there have been numerous concerns about the consequences of COMPAS scores on the decision making of judges  \citep{forrest2021machines}. In our view, the belief that the humans should be informed with the scores of a particular prediction problem is a very strong prior on the policy space of the machine: the policy space consists of a single policy. In order to credibly identify successful forms of human-machine cooperation, we should however consider a variety of plausible machine policies, and also account for the fact that human decision makers have to learn how to interact with them. This is exactly the setting that we consider in this paper. 

{\bf Human learning model.} In our model, there are no constraints on the algorithm that the two players might run. We also remain entirely agnostic about the policy spaces of both players. This serve the purpose of generality and keeps our work closely aligned with the extant literature on contextual bandits. However, these two assumptions are also major simplifications, especially insofar as the human decision maker is concerned. Indeed, it is a well-known fact that humans are not perfectly rational decision makers and have problems to deal with probabilities \citep{gigerenzer2005fast,kahneman2005model}. A human would not be able to correctly perform the updates prescribed by \Algorithm, MOSS, or any other bandit algorithm for that matter. The results provided in this paper apply to two perfectly rational decision makers who have access to arbitrary computational and cognitive resources. Two decision makers who only have access to limited computational and cognitive resources might hope to achieve as much, but will in general not be able to do any better. Of course it is an interesting question to ask how specific behavioral assumptions on the the human decision maker, such as bounded rationality \citep{selten1990bounded} or biases when dealing with machine recommendations \citep{green2019principles} influence optimal interaction. In the context of our model, such assumptions might take the form of assumptions on the policy space of the human, or the way in which the second decision maker selects policies in every round. This might be an interesting avenue for future research. Note, however, that in the context of our model, ``the human'' does not necessarily correspond to a single (biological) human. In most applications that we are interested in (compare (1)-(3) in Section \ref{sec:introduction}), there are many different judges or doctors that interact with a given machine learning system. While these human decision makers certainly learn individually how to interpret machine recommendations, they also engage in a collective learning procedure \citep{rakoff21}. While questions around the correct modelling of human-machine interaction are certainly very interesting, our objective in this paper is not to propose a universal model of human-machine interaction. Instead, our objective is to propose a model that is as simple as possible while still being able to capture the relations that we are interested in.

{\bf Exploration in high-stakes decision making problems.} In many human-machine decision making problems, direct exploration is highly problematic (for example in medical applications). In these applications, it is often impossible to explore according to an online algorithm during deployment. Instead, exploration is only possible during certain development stages (e.g. when we evaluate in a controlled study how doctors respond to different kinds of machine recommendations). In bandit models in particular, there are a number of different approaches  -- such as batching and offline learning -- that can be taken in order to model constraints on exploration \citep{amani2019linear,liu2020provably}. In any case, full online learning, that is the modelling approach taken in this paper, can only serve as a simple theoretical model for the process in which algorithmic decision aids are developed, tested and refined in practice (compare also Supplement \ref{apx:online_learning}).

{\bf Ethical impact.} This work discusses statistical efficiency, which is in itself not a sufficient criterion to justify automation. This is especially true in medicine, an area that is believed to experience the widespread deployment of machine learning systems in the future \citep{froomkin2019ais, Grote2020}. Automated decision making may also arise in undesired contexts. However, it remains important to understand it in the scenarios where it is desirable. As our work concerns theoretical foundations, theorems and proofs, we do not believe that it will have immediate negative consequences.

\subsubsection*{Acknowledgements}

We would like to thank S\'{e}bastien Bubeck, Nicol\`{o} Cesa-Bianchi and Thomas Grote for helpful discussions. We would also like to thank Ronja M{\"u}ller and Ruben Thoms for helping to create Figure 1. This work has been supported by the German Research Foundation through the Cluster of Excellence
“Machine Learning – New Perspectives for Science" (EXC 2064/1 number
390727645), the BMBF T\"ubingen AI Center (FKZ: 01IS18039A),  and the International Max Planck Research School for Intelligent Systems (IMPRS-IS).\\[25pt]

\bibliography{literature}

\begin{thebibliography}{65}
\providecommand{\natexlab}[1]{#1}
\providecommand{\url}[1]{\texttt{#1}}
\expandafter\ifx\csname urlstyle\endcsname\relax
  \providecommand{\doi}[1]{doi: #1}\else
  \providecommand{\doi}{doi: \begingroup \urlstyle{rm}\Url}\fi

\bibitem[Adebayo et~al.(2018)Adebayo, Gilmer, Muelly, Goodfellow, Hardt, and
  Kim]{adebayo2018sanity}
J.~Adebayo, J.~Gilmer, M.~Muelly, I.~Goodfellow, M.~Hardt, and B.~Kim.
\newblock Sanity checks for saliency maps.
\newblock In \emph{Advances in Neural Information Processing Systems}, 2018.

\bibitem[Amani et~al.(2019)Amani, Alizadeh, and Thrampoulidis]{amani2019linear}
S.~Amani, M.~Alizadeh, and C.~Thrampoulidis.
\newblock Linear stochastic bandits under safety constraints.
\newblock \emph{Advances in Neural Information Processing Systems}, 2019.

\bibitem[Angrist and Pischke(2008)]{angrist2008mostly}
J.~D. Angrist and J.-S. Pischke.
\newblock \emph{Mostly harmless econometrics}.
\newblock Princeton University Press, 2008.

\bibitem[Angwin et~al.(2016)Angwin, Larson, Mattu, and
  Kirchner]{angwin2016machine}
J.~Angwin, J.~Larson, S.~Mattu, and L.~Kirchner.
\newblock Machine bias. {ProPublica}.
\newblock \emph{See https://www. propublica.
  org/article/machine-bias-risk-assessments-in-criminal-sentencing}, 2016.

\bibitem[Audibert and Bubeck(2010)]{audibert2010regret}
J.-Y. Audibert and S.~Bubeck.
\newblock Regret bounds and minimax policies under partial monitoring.
\newblock \emph{The Journal of Machine Learning Research}, 2010.

\bibitem[Auer et~al.(2002)Auer, Cesa-Bianchi, Freund, and
  Schapire]{auer2002nonstochastic}
P.~Auer, N.~Cesa-Bianchi, Y.~Freund, and R.~E. Schapire.
\newblock The nonstochastic multiarmed bandit problem.
\newblock \emph{SIAM Journal on Computing}, 2002.

\bibitem[Avner and Mannor(2016)]{avner2016multi}
O.~Avner and S.~Mannor.
\newblock Multi-user lax communications: a multi-armed bandit approach.
\newblock In \emph{IEEE International Conference on Computer Communications},
  2016.

\bibitem[Baker et~al.(2020)Baker, Xiang, and Atkinson]{baker2020continuous}
S.~Baker, W.~Xiang, and I.~Atkinson.
\newblock Continuous and automatic mortality risk prediction using vital signs
  in the intensive care unit: a hybrid neural network approach.
\newblock \emph{Scientific Reports}, 2020.

\bibitem[Barocas et~al.(2019)Barocas, Hardt, and
  Narayanan]{barocas-hardt-narayanan}
S.~Barocas, M.~Hardt, and A.~Narayanan.
\newblock \emph{Fairness and Machine Learning}.
\newblock fairmlbook.org, 2019.
\newblock \url{http://www.fairmlbook.org}.

\bibitem[Beede et~al.(2020)Beede, Baylor, Hersch, Iurchenko, Wilcox,
  Ruamviboonsuk, and Vardoulakis]{beede2020human}
E.~Beede, E.~Baylor, F.~Hersch, A.~Iurchenko, L.~Wilcox, P.~Ruamviboonsuk, and
  L.~M. Vardoulakis.
\newblock A human-centered evaluation of a deep learning system deployed in
  clinics for the detection of diabetic retinopathy.
\newblock In \emph{CHI Conference on Human Factors in Computing Systems}, 2020.

\bibitem[Beygelzimer et~al.(2010)Beygelzimer, Langford, Li, Reyzin, and
  Schapire]{Beygelzimer2010ContextualBA}
A.~Beygelzimer, J.~Langford, L.~Li, L.~Reyzin, and R.~E. Schapire.
\newblock Contextual bandit algorithms with supervised learning guarantees.
\newblock In \emph{International Conference on Artificial Intelligence and
  Statistics}, 2010.

\bibitem[{\relax Board of Governors}(2007)]{reserve2007report}
{\relax Board of Governors}.
\newblock \emph{Report to the Congress on Credit Scoring and its Effects on the
  Availability and Affordability of Credit}.
\newblock Board of Governors of the US Federal Reserve System, 2007.

\bibitem[Boursier and Perchet(2019)]{boursier2019sic}
E.~Boursier and V.~Perchet.
\newblock {SIC-MMAB}: {s}ynchronisation involves communication in multiplayer
  multi-armed bandits.
\newblock In \emph{Advances in Neural Information Processing Systems}, 2019.

\bibitem[Bubeck et~al.(2020)Bubeck, Li, Peres, and Sellke]{bubeck2020non}
S.~Bubeck, Y.~Li, Y.~Peres, and M.~Sellke.
\newblock Non-stochastic multi-player multi-armed bandits: Optimal rate with
  collision information, sublinear without.
\newblock In \emph{Conference on Learning Theory}, 2020.

\bibitem[Carroll et~al.(2019)Carroll, Shah, Ho, Griffiths, Seshia, Abbeel, and
  Dragan]{NIPS2019_8760}
M.~Carroll, R.~Shah, M.~K. Ho, T.~Griffiths, S.~Seshia, P.~Abbeel, and
  A.~Dragan.
\newblock {On the Utility of Learning about Humans for Human-AI Coordination}.
\newblock In \emph{Advances in Neural Information Processing Systems}, 2019.

\bibitem[Cesa-Bianchi and Lugosi(2006)]{cesa2006prediction}
N.~Cesa-Bianchi and G.~Lugosi.
\newblock \emph{Prediction, Learning, and Games}.
\newblock Cambridge University Press, 2006.

\bibitem[De et~al.(2020{\natexlab{a}})De, Koley, Ganguly, and
  Gomez-Rodriguez]{de2020regression}
A.~De, P.~Koley, N.~Ganguly, and M.~Gomez-Rodriguez.
\newblock Regression under human assistance.
\newblock In \emph{AAAI}, 2020{\natexlab{a}}.

\bibitem[De et~al.(2020{\natexlab{b}})De, Okati, Zarezade, and
  Gomez-Rodriguez]{de2020classification}
A.~De, N.~Okati, A.~Zarezade, and M.~Gomez-Rodriguez.
\newblock Classification under human assistance.
\newblock \emph{arXiv preprint arXiv:2006.11845}, 2020{\natexlab{b}}.

\bibitem[De-Arteaga et~al.(2020)De-Arteaga, Fogliato, and
  Chouldechova]{de2020case}
M.~De-Arteaga, R.~Fogliato, and A.~Chouldechova.
\newblock A case for humans-in-the-loop: Decisions in the presence of erroneous
  algorithmic scores.
\newblock In \emph{CHI Conference on Human Factors in Computing Systems}, 2020.

\bibitem[Dietvorst et~al.(2015)Dietvorst, Simmons, and
  Massey]{dietvorst2015algorithm}
B.~J. Dietvorst, J.~P. Simmons, and C.~Massey.
\newblock Algorithm aversion: People erroneously avoid algorithms after seeing
  them err.
\newblock \emph{Journal of Experimental Psychology: General}, 2015.

\bibitem[Forrest(2021)]{forrest2021machines}
K.~B. Forrest.
\newblock \emph{When Machines Can Be Judge, Jury, and Executioner: Justice in
  the Age of Artificial Intelligence}.
\newblock World Scientific, 2021.

\bibitem[Froomkin et~al.(2019)Froomkin, Kerr, and Pineau]{froomkin2019ais}
A.~M. Froomkin, I.~Kerr, and J.~Pineau.
\newblock When {AIs} outperform doctors: confronting the challenges of a
  tort-induced over-reliance on machine learning.
\newblock \emph{Ariz. L. Rev.}, 61:\penalty0 33, 2019.

\bibitem[Gigerenzer and Kurzenhaeuser(2005)]{gigerenzer2005fast}
G.~Gigerenzer and S.~Kurzenhaeuser.
\newblock Fast and frugal heuristics in medical decision making.
\newblock In \emph{Science and medicine in dialogue: Thinking through
  particulars and universals}. Praeger Westport, CT, 2005.

\bibitem[Goldenberg and Engelhardt(2019)]{goldenberg2019health}
A.~Goldenberg and B.~Engelhardt.
\newblock Machine learning for computational biology and health.
\newblock Tutorial at Advances in Neural Information Processing Systems
  Conference, 2019.

\bibitem[Goodrich and Schultz(2007)]{Goodrich2007}
M.~Goodrich and A.~Schultz.
\newblock Human-robot interaction: A survey.
\newblock \emph{Foundations and Trends in Human-Computer Interaction}, 2007.

\bibitem[Green and Chen(2019)]{green2019principles}
B.~Green and Y.~Chen.
\newblock The principles and limits of algorithm-in-the-loop decision making.
\newblock \emph{Proceedings of the ACM on Human-Computer Interaction}, 2019.

\bibitem[Grote and Berens(2020)]{Grote2020}
T.~Grote and P.~Berens.
\newblock On the ethics of algorithmic decision-making in healthcare.
\newblock \emph{Journal of Medical Ethics}, 2020.

\bibitem[Hilgard et~al.(2019)Hilgard, Rosenfeld, Banaji, Cao, and
  Parkes]{hilgard2019learning}
S.~Hilgard, N.~Rosenfeld, M.~R. Banaji, J.~Cao, and D.~C. Parkes.
\newblock Learning representations by humans, for humans.
\newblock \emph{arXiv preprint arXiv:1905.12686}, 2019.

\bibitem[Ilyas et~al.(2019)Ilyas, Santurkar, Tsipras, Engstrom, Tran, and
  Madry]{NEURIPS2019_e2c420d9}
A.~Ilyas, S.~Santurkar, D.~Tsipras, L.~Engstrom, B.~Tran, and A.~Madry.
\newblock Adversarial examples are not bugs, they are features.
\newblock In \emph{Advances in Neural Information Processing Systems}, 2019.

\bibitem[Kahneman and Frederick(2005)]{kahneman2005model}
D.~Kahneman and S.~Frederick.
\newblock \emph{A model of heuristic judgment.}
\newblock Cambridge University Press, 2005.

\bibitem[Kalathil et~al.(2014)Kalathil, Nayyar, and
  Jain]{kalathil2014decentralized}
D.~Kalathil, N.~Nayyar, and R.~Jain.
\newblock Decentralized learning for multiplayer multiarmed bandits.
\newblock \emph{IEEE Transactions on Information Theory}, 2014.

\bibitem[Kleinberg et~al.(2018)Kleinberg, Lakkaraju, Leskovec, Ludwig, and
  Mullainathan]{kleinberg2018human}
J.~Kleinberg, H.~Lakkaraju, J.~Leskovec, J.~Ludwig, and S.~Mullainathan.
\newblock Human decisions and machine predictions.
\newblock \emph{The Quarterly Journal of Economics}, 2018.

\bibitem[Kleinberg et~al.(2019)Kleinberg, Ludwig, Mullainathan, and
  Sunstein]{kleinberg2018discrimination}
J.~Kleinberg, J.~Ludwig, S.~Mullainathan, and C.~R. Sunstein.
\newblock {Discrimination in the Age of Algorithms}.
\newblock \emph{Journal of Legal Analysis}, 10:\penalty0 113--174, 04 2019.

\bibitem[Lakhani and Sundaram(2017)]{lakhani2017deep}
P.~Lakhani and B.~Sundaram.
\newblock Deep learning at chest radiography: {A}utomated classification of
  pulmonary tuberculosis by using convolutional neural networks.
\newblock \emph{Radiology}, 2017.

\bibitem[Lakkaraju et~al.(2017)Lakkaraju, Kleinberg, Leskovec, Ludwig, and
  Mullainathan]{lakkaraju2017selective}
H.~Lakkaraju, J.~Kleinberg, J.~Leskovec, J.~Ludwig, and S.~Mullainathan.
\newblock The selective labels problem: Evaluating algorithmic predictions in
  the presence of unobservables.
\newblock In \emph{ACM SIGKDD International Conference on Knowledge Discovery
  and Data Mining}, 2017.

\bibitem[Lattimore and Szepesvari(2019)]{lattimore2019bandit}
T.~Lattimore and C.~Szepesvari.
\newblock \emph{Bandit Algorithms}.
\newblock Cambridge University Press, 2019.

\bibitem[Leonelli(2020)]{sep-science-big-data}
S.~Leonelli.
\newblock Scientific research and big data.
\newblock In E.~N. Zalta, editor, \emph{The Stanford Encyclopedia of
  Philosophy}. Metaphysics Research Lab, Stanford University, {S}ummer 2020
  edition, 2020.

\bibitem[Liu et~al.(2020)Liu, Swaminathan, Agarwal, and
  Brunskill]{liu2020provably}
Y.~Liu, A.~Swaminathan, A.~Agarwal, and E.~Brunskill.
\newblock Provably good batch reinforcement learning without great exploration.
\newblock \emph{arXiv preprint arXiv:2007.08202}, 2020.

\bibitem[Lucic et~al.(2020)Lucic, Haned, and de~Rijke]{lucic2020does}
A.~Lucic, H.~Haned, and M.~de~Rijke.
\newblock Why does my model fail? {C}ontrastive local explanations for retail
  forecasting.
\newblock In \emph{ACM Conference on Fairness, Accountability, and
  Transparency}, 2020.

\bibitem[Madras et~al.(2018)Madras, Pitassi, and Zemel]{madras2018predict}
D.~Madras, T.~Pitassi, and R.~S. Zemel.
\newblock Predict responsibly: Improving fairness and accuracy by learning to
  defer.
\newblock In \emph{Advances in Neural Information Processing Systems}, 2018.

\bibitem[Makino et~al.(2020)Makino, Jastrzebski, Oleszkiewicz, Chacko,
  Ehrenpreis, Samreen, Chhor, Kim, Lee, Pysarenko,
  et~al.]{makino2020differences}
T.~Makino, S.~Jastrzebski, W.~Oleszkiewicz, C.~Chacko, R.~Ehrenpreis,
  N.~Samreen, C.~Chhor, E.~Kim, J.~Lee, K.~Pysarenko, et~al.
\newblock Differences between human and machine perception in medical
  diagnosis.
\newblock \emph{arXiv preprint arXiv:2011.14036}, 2020.

\bibitem[Mart\'{\i}nez-Rubio et~al.(2019)Mart\'{\i}nez-Rubio, Kanade, and
  Rebeschini]{NEURIPS2019_85353d3b}
D.~Mart\'{\i}nez-Rubio, V.~Kanade, and P.~Rebeschini.
\newblock Decentralized cooperative stochastic bandits.
\newblock In \emph{Advances in Neural Information Processing Systems}, 2019.

\bibitem[Mas-Colell et~al.(1995)Mas-Colell, Whinston, Green,
  et~al.]{mas1995microeconomic}
A.~Mas-Colell, M.~D. Whinston, J.~R. Green, et~al.
\newblock \emph{Microeconomic theory}.
\newblock Oxford University Press, New York, 1995.

\bibitem[Mozannar and Sontag(2020)]{mozannar2020consistent}
H.~Mozannar and D.~Sontag.
\newblock Consistent estimators for learning to defer to an expert.
\newblock \emph{arXiv preprint arXiv:2006.01862}, 2020.

\bibitem[Neu(2015)]{neu2015.explore.no.more}
G.~Neu.
\newblock Explore no more: Improved high-probability regret bounds for
  non-stochastic bandits.
\newblock In \emph{Advances in Neural Information Processing Systems}, 2015.

\bibitem[Parasuraman et~al.(2000)Parasuraman, Sheridan, and
  Wickens]{parasuraman2000model}
R.~Parasuraman, T.~B. Sheridan, and C.~D. Wickens.
\newblock A model for types and levels of human interaction with automation.
\newblock \emph{IEEE Transactions on systems, man, and cybernetics-Part A:
  Systems and Humans}, 2000.

\bibitem[Patel et~al.(2019)Patel, Rosenberg, Willcox, Baltaxe, Lyons, Irvin,
  Rajpurkar, Amrhein, Gupta, Halabi, Langlotz, Lo, Mammarappallil, Mariano,
  Riley, Seekins, Shen, Zucker, and Lungren]{Patel_2019}
B.~N. Patel, L.~Rosenberg, G.~Willcox, D.~Baltaxe, M.~Lyons, J.~Irvin,
  P.~Rajpurkar, T.~Amrhein, R.~Gupta, S.~Halabi, C.~Langlotz, E.~Lo,
  J.~Mammarappallil, A.~J. Mariano, G.~Riley, J.~Seekins, L.~Shen, E.~Zucker,
  and M.~P. Lungren.
\newblock Human--machine partnership with artificial intelligence for chest
  radiograph diagnosis.
\newblock \emph{npj Digital Medicine}, 2019.

\bibitem[Pradier et~al.(2021)Pradier, Zazo, Parbhoo, Perlis, Zazzi, and
  Doshi-Velez]{pradier2021preferential}
M.~F. Pradier, J.~Zazo, S.~Parbhoo, R.~H. Perlis, M.~Zazzi, and F.~Doshi-Velez.
\newblock Preferential mixture-of-experts: Interpretable models that rely on
  human expertise as much as possible.
\newblock \emph{arXiv preprint arXiv:2101.05360}, 2021.

\bibitem[Raghu et~al.(2019)Raghu, Blumer, Corrado, Kleinberg, Obermeyer, and
  Mullainathan]{raghu2019algorithmic}
M.~Raghu, K.~Blumer, G.~Corrado, J.~Kleinberg, Z.~Obermeyer, and
  S.~Mullainathan.
\newblock The algorithmic automation problem: Prediction, triage, and human
  effort.
\newblock \emph{arXiv preprint arXiv:1903.12220}, 2019.

\bibitem[Rahwan et~al.(2019)Rahwan, Cebrian, Obradovich, Bongard, Bonnefon,
  Breazeal, Crandall, Christakis, Couzin, Jackson, et~al.]{rahwan2019machine}
I.~Rahwan, M.~Cebrian, N.~Obradovich, J.~Bongard, J.-F. Bonnefon, C.~Breazeal,
  J.~W. Crandall, N.~A. Christakis, I.~D. Couzin, M.~O. Jackson, et~al.
\newblock Machine behaviour.
\newblock \emph{Nature}, 2019.

\bibitem[Rakoff(2021)]{rakoff21}
J.~S. Rakoff.
\newblock Sentenced by algorithm.
\newblock \emph{The New York Review of Books}, 2021.

\bibitem[Rudin(2019)]{rudin2019stop}
C.~Rudin.
\newblock Stop explaining black box machine learning models for high stakes
  decisions and use interpretable models instead.
\newblock \emph{Nature Machine Intelligence}, 2019.

\bibitem[Sankararaman et~al.(2021)Sankararaman, Basu, and
  Sankararaman]{sankararaman2021dominate}
A.~Sankararaman, S.~Basu, and K.~A. Sankararaman.
\newblock Dominate or delete: Decentralized competing bandits in serial
  dictatorship.
\newblock In \emph{International Conference on Artificial Intelligence and
  Statistics}, 2021.

\bibitem[Sayres et~al.(2019)Sayres, Taly, Rahimy, Blumer, Coz, Hammel, Krause,
  Narayanaswamy, Rastegar, Wu, et~al.]{sayres2019using}
R.~Sayres, A.~Taly, E.~Rahimy, K.~Blumer, D.~Coz, N.~Hammel, J.~Krause,
  A.~Narayanaswamy, Z.~Rastegar, D.~Wu, et~al.
\newblock Using a deep learning algorithm and integrated gradients explanation
  to assist grading for diabetic retinopathy.
\newblock \emph{Ophthalmology}, 2019.

\bibitem[Seldin and Lugosi(2016)]{seldin2016lower}
Y.~Seldin and G.~Lugosi.
\newblock A lower bound for multi-armed bandits with expert advice.
\newblock In \emph{13th European Workshop on Reinforcement Learning (EWRL)},
  2016.

\bibitem[Selten(1990)]{selten1990bounded}
R.~Selten.
\newblock Bounded rationality.
\newblock \emph{Journal of Institutional and Theoretical Economics}, 1990.

\bibitem[Selvaraju et~al.(2017)Selvaraju, Cogswell, Das, Vedantam, Parikh, and
  Batra]{selvaraju2017grad}
R.~R. Selvaraju, M.~Cogswell, A.~Das, R.~Vedantam, D.~Parikh, and D.~Batra.
\newblock Grad-cam: Visual explanations from deep networks via gradient-based
  localization.
\newblock In \emph{International Conference on Computer Vision}, 2017.

\bibitem[Shamout et~al.(2020)Shamout, Zhu, Sharma, Watkinson, and
  Clifton]{8844833}
F.~E. Shamout, T.~Zhu, P.~Sharma, P.~J. Watkinson, and D.~A. Clifton.
\newblock Deep interpretable early warning system for the detection of clinical
  deterioration.
\newblock \emph{IEEE Journal of Biomedical and Health Informatics}, 2020.

\bibitem[Sheridan and Verplank(1978)]{sheridan1978human}
T.~B. Sheridan and W.~L. Verplank.
\newblock Human and computer controlof undersea teleoperators.
\newblock Technical report, MIT Man-Machine Systems Laboratory, Cambridge, MA,
  1978.

\bibitem[Simonyan et~al.(2014)Simonyan, Vedaldi, and
  Zisserman]{simonyan2013deep}
K.~Simonyan, A.~Vedaldi, and A.~Zisserman.
\newblock Deep inside convolutional networks: Visualising image classification
  models and saliency maps.
\newblock In \emph{ICLR Workshop}, 2014.

\bibitem[Tonekaboni et~al.(2018)Tonekaboni, Mazwi, Laussen, Eytan, Greer,
  Goodfellow, Goodwin, Brudno, and Goldenberg]{tonekaboni2018prediction}
S.~Tonekaboni, M.~Mazwi, P.~Laussen, D.~Eytan, R.~Greer, S.~D. Goodfellow,
  A.~Goodwin, M.~Brudno, and A.~Goldenberg.
\newblock {Prediction of Cardiac Arrest from Physiological Signals in the
  Pediatric ICU}.
\newblock In \emph{Machine Learning for Healthcare Conference}, 2018.

\bibitem[Tonekaboni et~al.(2019)Tonekaboni, Joshi, McCradden, and
  Goldenberg]{tonekaboni2019clinicians}
S.~Tonekaboni, S.~Joshi, M.~D. McCradden, and A.~Goldenberg.
\newblock What clinicians want: {C}ontextualizing explainable machine learning
  for clinical end use.
\newblock \emph{arXiv preprint arXiv:1905.05134}, 2019.

\bibitem[Tschandl et~al.(2020)Tschandl, Rinner, Apalla, Argenziano, Codella,
  Halpern, Janda, Lallas, Longo, Malvehy, et~al.]{tschandl2020human}
P.~Tschandl, C.~Rinner, Z.~Apalla, G.~Argenziano, N.~Codella, A.~Halpern,
  M.~Janda, A.~Lallas, C.~Longo, J.~Malvehy, et~al.
\newblock Human--computer collaboration for skin cancer recognition.
\newblock \emph{Nature Medicine}, 2020.

\bibitem[Von~Neumann and Morgenstern(2007)]{von2007theory}
J.~Von~Neumann and O.~Morgenstern.
\newblock \emph{Theory of games and economic behavior}.
\newblock Princeton University Press, 2007.

\bibitem[Wilder et~al.(2020)Wilder, Horvitz, and Kamar]{wilder2020learning}
B.~Wilder, E.~Horvitz, and E.~Kamar.
\newblock Learning to complement humans.
\newblock \emph{arXiv preprint arXiv:2005.00582}, 2020.

\end{thebibliography}

\clearpage
\appendix

\thispagestyle{empty}

\onecolumn \makesupplementtitle

\section{Proofs of theorems in the main paper}

\subsection{Additional definitions}  \label{apx:defintions}

Let $H_{1,t}\in (\mathcal{X}\times \mathcal{R}\times A\times [0,1])^t$ and $H_{2,t}\in (\mathcal{R}\times \mathcal{Z}\times A\times [0,1])^t$ be the histories of Player~1 and Player 2, respectively. Let $\mathcal{D}(X)$ denote the space of probability distributions over a space $X$, and $\mathcal{F}(X)$ the set of all finite subsets of $X$. An algorithm $A$ is a pair $A=(A_1,A_2)$ of two collections of measurable functions $A_1=(A_{1,t})_{t=1}^T$ and $A_2=(A_{2,t})_{t=1}^T$.  For $t=1$, we have $A_{1,1}:\mathcal{F}(\mathcal{R}^\mathcal{X})\times\mathcal{X}\to\mathcal{D}(\mathcal{R})$ and $A_{2,1}:\mathcal{F}(\mathcal{A}^{\mathcal{R}\times\mathcal{Z}})\times\mathcal{R}\times\mathcal{Z}\to\mathcal{D}(\mathcal{A})$. For $t=2,\dots,T$, we have $A_{1,t}:\mathcal{F}(\mathcal{R}^\mathcal{X})\times \mathcal{X} \times H_{1,t-1} \to \mathcal{D}(\mathcal{R})$ and $A_{2,t}:\mathcal{F}(\mathcal{A}^{\mathcal{R}\times\mathcal{Z}})\times \mathcal{R}\times \mathcal{Z}\times H_{2,t-1}\to \mathcal{D}(\mathcal{A})$. In Section \ref{sec:upper_bound}, we allow Player 2 to tell Player 1 which policy to use. This means that there is an additional collection of measurable functions $(A_{3,t})_{t=1}^T$ with $A_{3,1}:\mathcal{F}(\mathcal{A}^{\mathcal{R}\times\mathcal{Z}})\to\mathcal{D}(\{1,\dots,N_1\})$ and $A_{3,t}:\mathcal{F}(\mathcal{A}^{\mathcal{R}\times\mathcal{Z}})\times H_{2,t-1}\to \mathcal{D}(\{1,\dots,N_1\})$ for $t=2,\dots,T$. These functions specify the (possibly randomized) policies that Player 2 tells Player 1 to use. $A_1$ consists of the fixed functions that implement the said policy choices for the first player. Additionally, the history of Player 2  and domain of functions in $A_2$ contain the policy that Player 1 was told to use.

\subsection{Proof of Proposition \ref{prop:no_private_no_opacity}}
\label{apx:prop_no_private_no_opacity}

\begin{proof}
Without private information and opacity, the two players can perform actions that are equivalent to EXP4 run on the joint policy space $\Pi$ (the EXP4 Algorithm is reproduced in Supplement Figure \ref{fig:apx_exp4}). Note that without opacity, both players have access to $\Pi$. Since $x_t=z_t$, they are also able to evaluate $\pi(x_t,z_t)$ for all $\pi\in\Pi$. Hence, a trivial solution would be that the second player ignores the recommendations made by the first player and simply performs EXP4. The result then follows from the standard analysis of EXP4  \citep[Theorem 18.1]{lattimore2019bandit}. A solution more in line with the interaction in our model would be that the first player recommends, in each round, $r_t$ according to $\mathbb{P}(r_t=r)=q_{tr}$ where
\begin{equation*}
q_{tr}=\sum_{i=1}^{N_1}\sum_{j=1}^{N_2} Q_{t,ij}1_{\{f_i(x_t)=r\}}.
\end{equation*}
Here $Q_{t}\in\mathbb{R}^{N_1\times N_2}$ is the matrix maintained by EXP4 as described in Supplement Figure \ref{fig:apx_exp4}. The second player would then choose $a_t$ according to $\mathbb{P}(a_t=k)=p_{tk}$ with\begin{equation*}
\quad p_{tk}=\frac{\sum_{i=1}^{N_1}\sum_{j=1}^{N_2} Q_{t,ij}1_{\{f_i(x_t)=r_t \land g_{j}(r_t, z_t)=k\}}}{q_{t,r_t}},
\end{equation*}
i.e. there is a policy $g_t\in\Pi_2$ s.t. $a_t=g_t(r_t,z_t)$, while the action is again chosen exactly as in EXP4.
\end{proof}

\subsection{Proof of Proposition \ref{proposition1}} \label{apx:proposition1}

\begin{proof}
    Both players privately label their policies from $0,\dots,N_1-1$ and $0,\dots,N_2-1$. Before the game starts, both players agree on a deterministic strategy for solving an $N$-armed stochastic bandit problem. In round $t$, where arm $0\leq i\leq N-1$ is to be pulled in the $N$-armed stochastic bandit problem, for $i = a\cdot N_2 + b$ with $0\leq b < N_2$, Player 1 plays policy $a$ and Player 2 plays policy $b$. Since a deterministic strategy determines the next arm to be pulled solely on the basis of past pulled arms and obtained rewards, both players know which of the $N$ arms is to be pulled in each round.
    Agreeing on MOSS (Minimax  Optimal  Strategy  in  the  Stochastic  case), a variant of UCB,  allows the two players to bound the minimax regret by $25\sqrt{TN}$ (\cite{audibert2010regret}, Theorem 24).
\end{proof}

\subsection{Proof of Theorem \ref{thm:lower_bound} } \label{apx:proof_lower}

\subsubsection{Proof with private information}
\begin{proof} 

The idea is to construct a decision problem where the first player has to solve an $N_1$-armed stochastic Bernoulli bandit. The result then follows from the lower bound for stochastic Bernoulli bandits (e.g. Exercise 15.4 in \citet{lattimore2019bandit}). Note that Player 2 has only a single policy, i.e. $\Pi_2=\{g\}$. Thus, the assumption that Player 2 only plays actions that are suggested by policies in $\Pi_2$ effectively fixes $A_2$, the algorithm of the second player. 

Let $(X_{1t},\dots,X_{N_1,t})\in\{0,1\}^{N_1}$ be the payoffs associated with an $N_1$-armed stochastic Bernoulli bandit in round $t$. By assumption $K=2$, so $A=\{1,2\}$. Player 1 does not need to receive any context, so let $\mathcal{X}=\{\emptyset\}$. Choose $\mathcal{R}=\{1,\dots,N_1\}$ and $\Pi_1=\{f_i|f_i=i,i=1,\dots,N_1\}$. That is Player 1 has $N_1$ policies, and policy $f_i$ constantly suggests recommendation $i$. In effect, recommendations and policies are really the same, namely the arms of a stochastic bandit. Let $\mathcal{Z}=\{0,1\}^{N_1}$ and $\Pi_2=\{g\}$ with $g(r,z)=1+z_r$. For simplicity, let the payoff of Action 1 be 0 in all rounds. Conversely, let the payoff Action 2 be 1 in all rounds. Let the context vector $z_t$ of Player 2 be given by the payoffs associated with the Bernoulli bandit, i.e. $z_t=(X_{1t},\dots,X_{N_1,t})$.

In round $t$, when arm $i\in\{1,\dots,N_1\}$ of the Bernoulli bandit has payoff $X_{it}$, Player 2 assigns recommendation $i$ to action $1+X_{it}$. This results in a reward of $X_{it}$. Thus, in round $t$, where Player 1 chooses recommendation $r_t\in\{1,\dots,N_1\}$, the observes reward is $X_{r_t,t}$. To sum up, in every round, Player 1 incurs the reward of one of the arms of the Bernoulli bandit, and this arm can be freely chosen by choosing the recommendation. Since $z_t$ is not observed by Player 1, the payoffs of all other arms of the Bernoulli bandit remain unknown. Every algorithm for Player 1 gives rise to an algorithm for stochastic Bernoulli bandits and vice-versa, and we obtain the lower bound.
\end{proof}

\subsubsection{Proof with opacity}

\begin{proof}
As above, let $A=\{1,2\}$ and $\mathcal{R}=\{1,\dots,N_1\}$. Let $(X_{1t},\dots,X_{N_1,t})\in\{0,1\}^{N_1}$ be the payoffs associated with an $N_1$-armed stochastic Bernoulli bandit in round $t$. Now, in every round, both players receive the same context vector $x\in\{1,\dots,M\}$. The recommendations of policies of Player 1 are as before and independent of the context vector, $\Pi_1=\{f_i|f_i=i,i=1,\dots,N_1\}$.

The important part is the policy of Player 2, which is based on a function $\hat g:\{1,\dots,M\}\to\{0,1\}^{N_1}$. Instead of obtaining the payoffs of the Bernoulli bandit directly as contextual information, Player 2 now uses the private function $\hat g$ to obtain these payoffs from $x$. Naturally, $\hat g$ is not known to Player 1. As above, the policy of Player 2 is given by $g(r,x)=1+\hat g(x)_r$ and action payoffs are fixed to 0 and 1.

Let the context vector be uniformly distributed over $\{1,\dots,M\}$. We have to make sure that the same context vectors do not appear too often, since otherwise the first player could start to infer the payoffs associated with them. By choosing $M$ large enough, context vectors up to time $T$ are unique with probability arbitrarily close to 1. 

We still have to specify how to choose $\hat g$ as a function from $\{1,\dots,M\}$ to $\{0,1\}^{N_1}$. For $N_1$ and $M$ fixed, there are only finitely many of these functions. In order to realize a single desired Bernoulli bandit, draw $\hat g$ according to the probability distribution $\mathcal{\hat D}$ given by
\begin{equation*}
    \mathbb{P}_{\mathcal{\hat D}}(\hat g)=\prod_{i=1}^M \mathbb{P}\Big((X_{11},\dots,X_{N_1,1})=\hat g(i)\Big).
\end{equation*}
In other words, for all $i=1,\dots,M$, the distribution of $\hat g(i)$ over $\{0,1\}^{N_1}$ is exactly that of the Bernoulli bandit. 

By the same argument as in the prove with unknown context, if $\hat g$ is drawn according to $\mathcal{\hat D}$, Player~1 has to solve the Bernoulli bandit given by $(X_{1t},\dots,X_{N_1,t})$. Now recall that the minimax regret is given by
\begin{equation*}
R_T =\inf_{A_1}\sup_{\mathcal{D}}\sup_{|\Pi_1|=N_1}\sup_{|\Pi_2|=1}\,\Reg_T.
\end{equation*}
In particular,
\begin{equation*}
\sup_{|\Pi_2|=1}\,\Reg_T\geq \sup_{\mathcal{\hat D}}\,\mathbb{E}_{\hat g\sim\mathcal{\hat D}}\Big[\Reg_T\Big],
\end{equation*}
which shows the lower bound in terms of the minimax expected regret for $N_1$-armed stochastic Bernoulli bandits.
\end{proof}

\begin{figure}[t]\centering
\fbox{\small\parbox{0.83\linewidth}{\textbf{EXP4}\\[2pt]
\textbf{Parameters:} $\eta>0$, $\gamma>0$\\
\textbf{Initialization:} Vector $Q_{1}\in [0,1]^N$ with $Q_{1i}=\frac{1}{N}$\\
\textbf{For each} $t=1,...,T$
\begin{itemize} 
	\item[1.] Receive context $x_t$
	\item[2.] Choose action $a_t$ according to $p_{tk}=\sum_{i=1}^N Q_{ti} 1_{\{\pi_i(x_t)=k\}}$.
	\item[3.] Receive reward $y_t$ and estimate $\hat y_{tk}=1-\frac{1_{\{a_t=k\}}}{p_{tk}+\gamma}(1-y_t)$
    \item[4.] Propagates rewards to experts $\hat Y_{ti}=\hat y_{t,\pi_i(x_t)}$
    \item[5.] Player 2 updates $Q_t$ using exponential weighting\begin{equation*}
    Q_{t+1,i}=\frac{\exp\left(\eta \hat Y_{ti}\right) Q_{ti}}{\sum_{j}\exp\left(\eta \hat Y_{tj}\right)Q_{tj}}
    \end{equation*}
    \end{itemize}}}
\caption{EXP4. Adapted from Algorithm 11 in \cite{lattimore2019bandit}.}
\label{fig:apx_exp4}
\end{figure}

\subsection{Proof of Theorem \ref{thm:upper_bound}} \label{apx:proof_upper}

\begin{proof}
Recall the EXP4 algorithm, reproduced in Supplement Figure \ref{fig:apx_exp4}. The idea of the proof is as follows. In \Algorithm, Player 2 maintains a probability distribution over the space of all policy combinations $\Pi$ and performs importance-weighted updates. Player 2 does not know the policy space and context of Player 1. Therefore, in every round, he only obtains information on policy combinations where $f_{i_t}$, the function that the first player actually played, is present. This restricts Player 2 and does not allow him to perform the sames updates as EXP4. However, assume that all policy combinations where $f_{i_t}$ is not present had suggested different actions than the policy combinations where $f_{i_t}$ is present. In this case, the updates in \Algorithm would be equivalent to the updates of EXP4. Therefore, we now construct a bandit problem where two different policies of Player 1 never suggest the same action, and show that Algorithm 1 is equivalent to EXP4 on this related bandit problem.

Consider the adversarial contextual bandit problem with $KN_1$ actions and policy space 
\begin{equation*}
\begin{split}
\tilde\Pi=\Big\{h_{i,j}\,\big|&i=1,\dots,N_1,\,j=1,\dots,N_2,\\ &h_{i,j}:\mathcal{X}\times\mathcal{Z}\to\{1,\dots,KN_1\},\\
& h_{i,j}(x,z)=(i-1)K+g_j(f_i(x),z)\Big\}.
\end{split}
\end{equation*} This policy space consists of $N$ policies, and there exists a natural bijection $I$ between $\tilde\Pi$ and $\Pi$ given by $ h_{i,j}\mapsto g_j(f_i(\cdot),\cdot)$.
Let the adversarial payoffs of this new problem be a function of the (adversarial or i.i.d.) payoffs of the original problem, namely
\begin{equation*}
\tilde x_t=(x_t,z_t)
\end{equation*}
and
\begin{equation*}
\tilde Y_t(k)= Y_t\Big(1+((k-1)\,\text{mod}\, K)\Big),
\end{equation*}for all $t=1,\dots,T$ and $k=1,\dots,KN_1$. Here $Y_t\in [0,1]^A$ contains the payoffs of the original problem, and $\tilde Y_t\in [0,1]^{\{1,\dots,KN_1\}}$ the payoffs of the new problem. By construction,
\begin{equation*}
\tilde Y_t\Big(h_{i,j}(\tilde x_t)\Big) = Y_t\Big(g_j(f_i(x_t),z_t)\Big).
\end{equation*} Therefore,
\begin{equation}\label{eq:proof_up_max}
\max_{\tilde\pi\in\tilde\Pi}\sum_{t=1}^T \tilde Y_t\Big(\tilde\pi(\tilde x_t)\Big)=\max_{\pi\in\Pi}\sum_{t=1}^T Y_t\Big(\pi(x_t,z_t)\Big).
\end{equation}We are now going to show that \Algorithm is equivalent to EXP4(-IX) on this adversarial contextual bandit problem. In this proof, we denote all variables related this problem and EXP4 with a $\sim$. For example, $\tilde a_t$ is the action chosen by EXP4 in round $t$, resulting in a payoff of $\tilde y_t$. Since both \Algorithm and EXP4 are randomized, equivalence means that there exists a coupling of the random variables drawn by both algorithms under which, in all rounds, the probability distribution $Q_t$ maintained by \Algorithm is the probability distribution $\tilde Q_t$ maintained by EXP4 (with respect to bijection $I$), $\tilde a_t=(i_t-1)K+a_t$ and $\tilde y_t=y_t$. 

We proceed by induction over $t$. The induction hypothesis is that equivalence holds up to round $t$. This is obviously true in the first round since both $Q_t$ and $\tilde Q_t$ are initialized to be uniform. In round $t$, EXP4 chooses an action $\tilde a_t\in\{1,\dots,KN_1\}$. This action $\tilde a_t$ can be uniquely written as $\tilde a_t=(\hat i_t-1) K+ \hat a_t$ for some $\hat i_t\in\{1,\dots,N_1\}$ and $\hat a_t\in\{1,\dots,K\}$. By construction, it is exactly policies $h_{i,1},\dots,h_{i,N_2}$ that suggest actions
\begin{equation*}
(i-1)K+1,\dots,iK.    
\end{equation*} Hence, 
\begin{equation*}
 \mathbb{P}\left(\hat i_{t}=i\right)=\sum_{j=1}^{N_2}Q_{t,ij}=\mathbb{P}(i_t=i),  
\end{equation*} 
where the first equality is due to the induction hypothesis and the second  due to the definition of $q_{ti}$ in \Algorithm. Since they have the same distribution, $\hat i_t$ and $i_t$ can be perfectly coupled. Additionally, and already subject to this coupling,
\begin{equation*}
\begin{split}
\mathbb{P}(\hat a_t=k\,|\,i_t=i)&=\frac{\mathbb{P}(\tilde a_t=(i-1)K+k)}{\mathbb{P}(i_t=i)}\\[4pt]
&=\frac{\sum_{j=1}^{N_2}Q_{t,ij}1_{\big\{h_{i,j}(\tilde x_t)=(i-1)K+k\big\}}}{\sum_{j=1}^{N_2}Q_{t,ij}}\\[4pt]
&=\mathbb{P}(a_t=k \,|\, i_t=i)
\end{split}
\end{equation*} where we used the definition of $a_t$ in Equation (\ref{eq:alg1_pt}) of the main paper and the fact that
\begin{equation*}
h_{i,j}(\tilde x_t)=(i-1)K+k\iff g_j(f_i(x_t),z_t)=k.    
\end{equation*} Thus, conditional on $i_t$, $\hat a_t$ and $a_t$ have the same distribution. Therefore, $\hat a_t$ and $a_t$ can be perfectly coupled, too, and we arrive at $\tilde a_t=(i_t-1)K+a_t$. From the definition of $\tilde Y_t$, it follows that $\tilde y_t=y_t$.

It remains to show that the update $Q_{t}\to Q_{t+1}$ in \Algorithm agrees with EXP4. We have to show that $\hat Y_t$ in \Algorithm agrees with the importance-weighted reward estimates of EXP4. We distinguish three cases. The first case is $i=i_t$ and $g_j(f_i(x_t),z_t)= a_t$. Here it holds that 
\begin{equation*}
\begin{split}
\hat Y_{t,ij}&=0+1-\frac{1}{q_{t,i_t}\,p_{t,a_t}+\gamma}(1-y_t)\\
&=1-\frac{1}{\tilde p_{tk} +\gamma}(1-\tilde y_t).    
\end{split}
\end{equation*}
The second case is $i=i_t$ and $g_j(f_i(x_t),z_t)\neq a_t$. Here it holds that $\hat Y_{t,ij}=0+1=1$. The third case is $i\neq i_t$. Here it holds that $\hat Y_{t,ij}=1+0=1$, too. In all three cases, the update agrees exactly with EXP4. 

We have shown that $\sum_{t=1}^{T}y_t=\sum_{t=1}^{T}\tilde y_t$. Subtracting this from \eqref{eq:proof_up_max}, we see that  
\begin{equation}
\max_{\pi\in\Pi}\sum_{t=1}^T Y_t\Big(\pi(x_t,z_t)\Big)-\sum_{t=1}^T  y_t = \max_{\tilde\pi\in\tilde\Pi}\sum_{t=1}^T \tilde Y_t\Big(\tilde\pi(\tilde x_t)\Big)-\sum_{t=1}^T \tilde y_t.
\end{equation}
From the analysis of EXP4, e.g. from Theorem 18.1 in \cite{lattimore2019bandit}, we know that
\begin{equation*}
\mathbb{E}\left(\max_{\tilde\pi\in\tilde\Pi}\sum_{t=1}^T\tilde\pi(\tilde x_t)-\sum_{t=1}^T \tilde y_t\right)\leq \sqrt{2TK N_1\ln(N_1 N_2)}
\end{equation*}for $\gamma=0$ and $\eta=\sqrt{2\log(N_1N_2)/(TKN_1)}$, which implies the desired bound.
\end{proof}

\subsection{Proof of Theorem \ref{thm:policy_space_independence}} \label{apx:proof_policy_space_independence}

\begin{proof} Assume that $|\mathcal{R}|<\infty$, otherwise the bound is vacuous. Let $f_1$ and $f_2$ be two policies of Player 1. In general, the expected regret under $f_1$ and $f_2$ depends on the policy choice of Player 2. Specifically, there might be $g_1$ and $g_2$ such that $Y\big(g_1(f_1(x),z)\big)>Y\big(g_1(f_2(x),z)\big)$ and $Y\big(g_2(f_1(x),z)\big)<Y\big(g_2(f_2(x),z)\big)$. Let $\pi_\star=(g_\star,f_\star)$ be an optimal policy combination. Under policy space independence, the quantities
\begin{equation*}
\Reg(f)=Y\big(g(f_\star(x),z)\big)-Y\big(g(f(x),z)\big)  
\end{equation*}
and 
\begin{equation*}
\Reg(g)=Y\big(g_\star(f(x),z)\big)-Y\big(g(f(x),z)\big)   
\end{equation*}
are well-defined. Moreover,
\begin{equation*}
Y\big(\pi_\star\big)-Y\big(g(f(x),z)\big)=\Reg(g)+\Reg(f).
\end{equation*}
That both players explore independently using EXP4 means the following. Player 2 uses EXP4 on $A$ with $\eta_1=\sqrt{2\log(N_2)/(TK)}$ and $\gamma_1=0$. Player~1 considers recommendations as actions and uses EXP4 on $\mathcal{R}$ with $\eta_2=\sqrt{2\log(N_1)/(T|\mathcal{R}|)}$ and $\gamma_2=0$. In round $t$, there exist policies $f_{i_t}$ and $g_{j_t}$ such that $r_t=f_{i_t}(x_t)$ and $a_t=g_{j_t}(f_{i_t}(x_t), z_t)$. Player 1 solves the adversarial contextual bandit problem with context $x_t$, action space $\mathcal{R}$ and policy space $\Pi_1$. Player 2 solves the adversarial contextual bandit problem with context $(r_t,z_t)$, action space $A$ and policy space $\Pi_2$. Player 1 provides adversarial context for Player 2, and Player 2 provides adversarial payoff for Player~1. Because of policy space independence, this independent exploration strategy also controls the joint expected regret. 

First note that $i_t$ and $j_t$ are functions of the history and can be considered drawn before the tuple $(x_t,z_t,Y_t)$. The expected regret in round $t$ is given by
\begin{equation*}
\mathbb{E}_{(x_t,z_t,Y_t)\sim\mathcal{D}}\Big[Y_t(g_\star(f_\star(x_t),z_t)- Y_t(g_{j_t}(f_{i_t}(x_t), z_t))\Big]=Y\big(g_\star(f_\star(x),z)\big)-Y\big(g_{j_t}(f_{i_t}(x), z)\big).
\end{equation*}
Making use of policy space independence, the right hand side can be rewritten as
\begin{equation*}
\begin{split}
& Y\big(g_\star(f_\star(x),z)\big)-Y\big(g_\star(f_{i_t}(x),z)\big)+Y\big(g_\star(f_{i_t}(x),z)\big)-Y\big(g_{j_t}(f_{i_t}(x), z)\big)\\[5pt]
=\,\,&Y\big(g_{j_t}(f_\star(x),z)\big)-Y\big(g_{j_t}(f_{i_t}(x),z)\big)+Y\big(g_\star(f_{i_t}(x),z)\big)-Y\big(g_{j_t}(f_{i_t}(x), z)\big).
\end{split}
\end{equation*}
Summing over $t$, the expected regret is given by
\begin{equation*}
\begin{split}
\Reg_T=&\sum_{t=1}^T\Big[Y\big(g_{j_t}(f_\star(x),z)\big)-Y\big(g_{j_t}(f_{i_t}(x),z)\big)\Big]\\
+&\sum_{t=1}^T\Big[Y\big(g_\star(f_{i_t}(x),z)\big)-Y\big(g_{j_t}(f_{i_t}(x), z)\big)\Big].
\end{split}
\end{equation*}
The first sum is the expected regret in the adversarial contextual bandit problem of the first player. The second sum is the expected regret in the adversarial contextual bandit problem of the second player. From the analysis of EXP4, e.g. from Theorem 18.1 in \cite{lattimore2019bandit}, we obtain
\begin{equation*}
\sum_{t=1}^T\Big[Y\big(g_{j_t}(f_\star(x),z)\big)-Y\big(g_{j_t}(f_{i_t}(x),z)\big)\Big]\leq\sqrt{2T|\mathcal{R}|\ln N_1} 
\end{equation*}
and
\begin{equation*}
\sum_{t=1}^T\Big[Y\big(g_\star(f_{i_t}(x),z)\big)-Y\big(g_{j_t}(f_{i_t}(x), z)\big)\Big]\leq\sqrt{2TK\ln N_2}   ,   
\end{equation*}
which implies the desired bound.
\end{proof}

\section{Fixed rules that allocate decisions result in policy space independence} \label{apx:fixed_allocation_rules}

In this section we formalize the example given in Section \ref{sec:fixed_allocation_rules}. We show that fixed rules that allocate decisions to either the human or the machine result in policy space independence. Let $\mathcal{R}=A$ (treatment recommendations), $D:\mathcal{Z}\mapsto\{0,1\}$ (the human decides who decides), $\tilde \Pi_2:\mathcal{Z}\to A$ (the human's own decision rules) and
\begin{equation*}
\Pi_2=\{g|g=D(z)\tilde g(z)+(1-D(z))r,\,\tilde g\in\tilde\Pi_2\}.    
\end{equation*}
Here $r=f(x)$ where $f\in\Pi_1$ is the decision rule used by the machine. Now, for all $f\in\Pi_1$ and $g\in\Pi_2$, and all distributions $\mathcal{D}$,
\begin{equation*}
\begin{split}
Y(g(f(x),z))&=\mathbb{E}_{(x,y,z)\sim\mathcal{D}}[Y(g(f(x),z))]\\[4pt]
&=\mathbb{P}(D(z)=0)\mathbb{E}[Y(g(f(x),z))|D(z)=0]\\[4pt]
&\quad+\mathbb{P}(D(z)=1)\mathbb{E}[Y(g(f(x),z))|D(z)=1]\\[4pt]
&=\mathbb{P}(D(z)=0)\mathbb{E}[Y(f(x))|D(z)=0]+\mathbb{P}(D(z)=1)\mathbb{E}[Y(\tilde g(z))|D(z)=1].\\[2pt]
\end{split}
\end{equation*} Thus,
\begin{equation*}
\begin{split}
Y(g_1(f_1(x),z)) - Y(g_1(f_2(x),z)) &= \mathbb{P}(D(z)=0)\mathbb{E}[Y(f_1(x))-Y(f_2(x))|D(z)=0]\\[4pt]
&=Y(g_2(f_1(x),z)) - Y(g_2(f_2(x),z))
\end{split}
\end{equation*}
for all $f_1,f_2\in\Pi_1$, $g_1,g_2\in\Pi_2$ and all distributions $\mathcal{D}$. In the key step of the derivation, we did not use the fact that $D$ was a (measurable) function of $Z$. Indeed, the sample space can be partitioned with respect to any event $D$.

\section{Fixed second player in Theorem \ref{thm:lower_bound}} 
\label{apx:lower_assumption_construction}

In Theorem \ref{thm:lower_bound}, we assumed that Player 2 only plays actions that are suggested by policies in $\Pi_2$. We are convinced that this assumption can be dropped if the problem instances in the respective proofs are modified in the following two ways. 

First, in every round, the relation between policies and recommendations should be entirely random. Concretely, let the policies of Player 1 depend on a context vector $x\in\{1,\dots,M\}$. In every round, let $x_t$ be uniform on $\{1,\dots,M\}$. Moreover, choose $M$ large enough such that every context vector occurs at most once up to time $T$. For every $x\in\{1,\dots,M\}$, randomly draw a permutation $\pi_x\in S_{N_1}$. Choose the policy space of the first player such that given context $x_t$, policy $f_i$ recommends $\pi_{x_t}(i)$. In effect, up to time $T$, the policies of Player 1 make random recommendations, subject to the constraint that all recommendations be different. 

Second, in every round, it should be entirely random which action gives the payoff of 1. Thus, for every $x\in\{1,\dots,M\}$, randomly drawn one action to give a payoff of 1, and set the payoff of the other action to 0.

In the first proof of Theorem \ref{thm:lower_bound} (unknown context), permute the context vector $z$ of Player 2 so that every policy still gets the same payoff as it would in the original construction (considering both $\pi_x$ and the  permuted payoffs). In the second proof of Theorem \ref{thm:lower_bound} (unknown policy), let the policy of Player 2 encode the appropriately permuted context vector.

Intuitively, if Player 2 knew which policies suggested which recommendations, Player 2 could effectively learn for Player 1. This is since Player 2 does always know the relation between recommendations and actions. In the given problem instance, the relation between policies and recommendations is impossible to know, at least up to time $T$.

\section{Problem instance for Conjecture \ref{thm:conjecture}} \label{apx:conjecture}

In this section we give a problem instance for Conjecture \ref{thm:conjecture}. We conjecture that it is a worst-case instance for which the lower bound stated in Conjecture \ref{thm:conjecture} holds. 

In every round, let one action give a payoff of 0 and the other a payoff of 1. Randomly decide in every round which action gives the payoff of 1. Choose the context vector and policy class of Player 1 such that he uniformly receives one of the  $2^{N_1}$ possible expert recommendations in every round. Ahead of time, select a policy of Player 1 and Player 2, respectively (the optimal policies). In every round, a policy for Player 2 gives a map $\mathcal{R}\to A$. With $\mathcal{R}=A=\{0,1\}$, there are 4 possible maps that we denote by $(0,0)$, $(1,0)$, $(0,1)$ and $(1,1)$. Here $(1,0)$ is the map that maps recommendation 0 to Action 1 and recommendation 1 to action 0. For the optimal policy of Player 2, let the relation between recommendations and actions be such that every policy of Player 1 except the optimal policy receives an expected payoff of 0.5, and the optimal policy receives an expected payoff of $0.5+\Delta$. This can be achieved as follows. In round $t$, where the optimal policy recommends $r_t$, map recommendation $r_t$ to the action with a payoff of 1 with probability $0.5+\Delta$. Similarly, map recommendation $1-r_t$ to the action with a payoff of 1 with probability $0.5-\Delta$. Note that since context vectors of Player 1 are drawn uniformly at random, each policy makes the same recommendation as the optimal policy exactly half of the time. For all other policies of Player 2, draw one of the 4 possible maps from recommendations to actions according to 
\begin{equation*}
\begin{split}
&\mathbb{P}\big((0,0)\big)=0.25-\Delta^2,\qquad\mathbb{P}\big((1,0)\big)=0.25+\Delta^2\\
&\mathbb{P}\big((0,1)\big)=0.25+\Delta^2,\qquad\mathbb{P}\big((1,1)\big)=0.25-\Delta^2.
\end{split}
\end{equation*}
This distribution is chosen such that all other policies have the same marginal distribution over the maps from recommendations to actions as the optimal policy. 

Let us quickly outline why we think that this is a difficult problem instance. Imagine that in every round, both players choose a policy according to some decision rule. If both players choose their optimal policy, the expected payoff is 
$0.5+\Delta$. Should any of the two players not choose their optimal policy, the expected payoff is $0.5$ (for all policy choices of the other player, also the optimal policy). Now consider what happens in the first round of the game. Assume that both players choose a policy uniformly at random (uniformly choosing recommendations, maps or actions does not reveal any information at all). Then, the expected payoff of the optimal policies of both players is $0.5+\frac{\Delta}{N_1}$. Thus, at least in the first round, the magnitude of the signal is $\frac{\Delta}{N_1}$, while the magnitude of the regret is $\Delta$. While the magnitude of the signal increases as the other player starts to identify the optimal policy, this strongly suggests that the regret does not scale logarithmically in $N_1$.

\section{Online learning and repeated supervised learning} \label{apx:online_learning}

In this section we give some more detail on why online learning is the correct approach to study human-machine decision making. Indeed, full online learning, as studied in our paper, is the most general and unrestricted way to understand how decisions evolve over time. This is despite the fact that machine learning algorithms are often not deployed in an online fashion. One reason for the latter is that online learning entails exploration which usually requires informed consent of the individuals who are impacted by the decisions. 

In practice, machine learning algorithms are usually trained on a historical dataset. In a human-machine decision making context, one would then evaluate how well humans perform with the trained algorithm, or a given number of trained algorithms. This might include some form of training for human decision makers plus a randomized controlled trial. If one finds that a given system performs sufficiently well, it might be deployed . Although this procedure is not an explicit online learning procedure, it is subject to the same limitations as online learning, at least insofar as coordination between the two decision makers is concerned. Viewed through the lens of our model, it could be interpreted as follows. First, the human makes a number of decisions, ignoring the machine (this produces the historical dataset). Second, the machine decides on a number of candidate policies (this is the supervised learning part). Third, the human tries to learn how to interpret the candidate policies of the machine (as in Section \ref{sec:upper_bound}). A slightly different interpretation would be to consider the result of supervised learning as the initial policy space of the machine. 
More generally, {\it full online learning is the  theoretical limit} of all sorts of procedures that iterate between machine learning on a given dataset, evaluating how well something works with humans in a real-world setting, collecting a bigger dataset, retraining our model in order to improve performance, evaluating again with humans, and so on. Importantly, online learning covers the scenario where we continuously collect data as a given system is running and then re-train it, say, once a year. In fact, full online learning places as few constraints on learning as possible. For example, re-training a system only at fixed intervals introduces an additional constraint often referred to as batching.

\section{Opacity and implicit communication} \label{apx:implicit_coordination}

In this section we discuss a theoretical subtlety that arises due to the way in which we set the problem up. This gives more details on the discussion at the end of Section \ref{sec:lower_bound} and in Supplement \ref{apx:lower_assumption_construction}.

We model opacity by keeping knowledge about the policy spaces to the respective players. As is apparent from the definition of the minimax regret in section \ref{sec:formal_setup}, both players first fix the way in which they want to approach the problem (the algorithm), then get to see the respective policy spaces. Importantly, we decided to place {\it no} restrictions on the algorithm that the two players might run. This is because the algorithm is part of the solution and not part of the problem. It also keeps our work closely aligned with the extant literature on online learning. This assumption has, however, a subtle consequence. Namely, the algorithms of both players can be arbitrarily well adapted. In a sense, before the game starts, the two players are allowed to get together in order to discuss how the problem might be approached. During the game, players might then try to implicitly encode information about policy spaces and context in actions and recommendations -- according to some protocol that they agreed upon in advance. 

With regard to our original research question, elaborate implicit communication protocols between the two players are of course unrealistic and even violate the idea of opacity. After all, it is implausible that a computer program and a human decision maker would communicate with such means. In this regard, note that we ruled out implicit communication protocols in Theorem \ref{thm:lower_bound} by assuming that the second player follows his one (and only) policy.

From a theoretical perspective, the question of whether implicit communication protocols would make a difference nevertheless remains interesting \citep{bubeck2020non}. As we argue in Supplement \ref{apx:lower_assumption_construction}, we believe that this is not the case.

\vfill

\end{document}